\documentclass[10pt,journal,cspaper,compsoc]{IEEEtran}
\ifCLASSOPTIONcompsoc
   \usepackage[nocompress]{cite}
\else
   \usepackage{cite}
\fi
\ifCLASSINFOpdf
   \usepackage[pdftex]{graphicx}
\else
\fi
\usepackage[cmex10]{amsmath}
\usepackage{bm,amssymb,graphicx}
\usepackage{algorithm}
\usepackage{algorithmic}
\usepackage{color}

\usepackage{array}

\usepackage{mdwmath}
\usepackage{mdwtab}

\ifCLASSOPTIONcompsoc
\usepackage[tight,normalsize,sf,SF]{subfigure}
\else
\usepackage[tight,footnotesize]{subfigure}
\fi
\usepackage{url}

\begin{document}
%
\title{Manifold Partition Discriminant Analysis}
%
%
%
%

\author{Yang Zhou and Shiliang~Sun
\IEEEcompsocitemizethanks{\IEEEcompsocthanksitem Yang Zhou and Shiliang Sun (corresponding author) are
with Shanghai Key Laboratory of Multidimensional Information Processing, Department of Computer Science and
Technology, East China Normal University, 500 Dongchuan Road, Shanghai  200241, P. R. China
(email: shiliangsun@gmail.com, slsun@cs.ecnu.edu.cn) \protect\\
}
\thanks{Manuscript received Dec. 21, 2014; revised Apr. 2, 2015, Sep. 10, 2015 and Jan. 21, 2016; accepted Feb. 9, 2016.}}

%
%

\graphicspath{{figures/}}

\markboth{IEEE TRANSACTIONS ON CYBERNETICS}%
{Shell \MakeLowercase{\textit{et al.}}: Bare Demo of IEEEtran.cls for Computer Society Journals}
%


\IEEEcompsoctitleabstractindextext{%
\begin{abstract}
We propose a novel algorithm for supervised dimensionality reduction
named Manifold Partition Discriminant Analysis (MPDA). It aims to
find a linear embedding space where the within-class similarity is
achieved along the direction that is consistent with the local
variation of the data manifold, while nearby data belonging to
different classes are well separated. By partitioning the data
manifold into a number of linear subspaces and utilizing the
first-order Taylor expansion, MPDA explicitly parameterizes the
connections of tangent spaces and represents the data manifold in a
piecewise manner. While graph Laplacian methods capture only the
pairwise interaction between data points, our method capture both
pairwise and higher order interactions (using regional consistency)
between data points. This manifold representation can help to
improve the measure of within-class similarity, which further leads
to improved performance of dimensionality reduction. Experimental
results on multiple real-world data sets demonstrate the
effectiveness of the proposed method.
\end{abstract}

}

\maketitle

\IEEEdisplaynotcompsoctitleabstractindextext

%
\IEEEpeerreviewmaketitle

\section{Introduction}

Linear Discriminant Analysis (LDA) is a classical supervised
dimensionality reduction method. It aims to find an optimal
low-dimensional projection along which data points from different
classes are far away from each other, while those belonging to the
same class are as close as possible. In the resultant
low-dimensional space, the performance of classifiers could be
improved. Because of this, LDA is especially useful for
classification tasks. Due to its effectiveness, LDA is widely
employed in different applications such as face recognition and
information retrieval
\cite{dai2007face,zhao2008incremental,Dornaika13el,Wang14FDA}.
However, when the input data are multimodal or mainly characterized
by their variances, LDA cannot perform very well. This is caused by
the assumption implicitly adopted by LDA that data points belonging
to each class are generated from multivariate Gaussian distributions
with the same covariance matrix but different means. If data are
formed by several separate clusters or lie on a manifold, this
assumption is violated, and thus LDA obtains undesired results.

To solve this problem, some extensions of LDA have been proposed,
which resort to discovering local data structures. Marginal Fisher
Analysis (MFA) \cite{yan2007graph} aims to gather the nearby
examples of the same class, and separate the marginal examples
belonging to different classes. Locality Sensitive Discriminant
Analysis (LSDA) \cite{cai2007locality} maps data points into a
subspace where the examples with the same label at each local area
are close, while the nearby examples from different classes are
apart from each other. Local Fisher Discriminant Analysis (LFDA)
\cite{sugiyama2007dimensionality} also focuses on discovering local
data structures. It can be viewed as performing LDA on the local
area around each data point. LFDA is a very effective algorithm and
has many applications. Recently, LFDA (combined with PCA) was
applied to the pedestrian re-identification problem and achieved the
state-of-the-art performance \cite{pedagadi2013local}. Despite of
different names and motivations, these methods, in fact, fall into
the same graph Laplacian based framework. All of them employ the
Laplacian matrix on specific graphs to characterize data structures
locally, and share the same idea that if nearby examples $x_i, x_j$
have the same class label $y$, they should be projected as close as
possible, otherwise, they should be well separated. By exploiting
the local structures around each data point, they are able to
process the data on which LDA cannot achieve reasonable results. As
widely recognized, graphs are often used as a proxy for the
manifold. Therefore, these methods, to some extent, can be viewed as
the combinations of manifold learning and LDA.

Although the above methods
overcome the drawback of LDA, they rely on
the graph Laplacian to capture the manifold structure,
where only pairwise differences are considered
whereas regional consistency
is ignored. The regional consistency can
be characterized by tangent spaces of the data manifold,
which could be very useful to enhance the
performance of discriminant analysis in some situations
\cite{NIPS1992_656} \cite{sun2013tangent}.
Moreover, the definition of closeness of these
graph Laplacian based methods is rather vague.
Along which direction can we decide if the closeness
of the mapped data points is achieved?
We advocate that
in order to preserve the manifold structure as much as possible,
the closeness of the embeddings should be achieved along the
direction that is consistent with the local variation
of the data manifold.

During recent years, tangent space based methods
have received considerable interest in the area of manifold learning
\cite{Zhang02principalmanifolds,brand2002charting,sun2013tangent,lin2013parallel}.
They utilize tangent spaces to estimate and extract the topological and
geometrical structure of the underlying manifold.
Local Tangent Space Alignment (LTSA) \cite{Zhang02principalmanifolds}
constructs tangent spaces at each data point and then
aligns them to obtain a global coordinate through
minimizing the reconstruction error.
Similar to LTSA, Manifold Charting \cite{brand2002charting}
tries to unfold the manifold by aligning local charts.
Tangent Space Intrinsic Manifold Regularization (TSIMR) \cite{sun2013tangent}
estimates a local linear function on the manifold which has constant
manifold derivatives. Parallel Vector Field Embedding (PFE)
\cite{lin2013parallel} represents a function along the manifold
from the perspective of vector fields and requires the vector
field at each data point to be as parallel as possible.
Due to exploiting the regional consistency reflected by tangent spaces,
these tangent space based methods work well for representing
the manifold structure. However, because of their unsupervised nature,
they have no ability to capture the
discriminative information from class labels, and
thus are not optimal for supervised dimensionality reduction.
Then how should we utilize the regional consistency of tangent spaces to
improve the performance of supervised dimensionality reduction?

Besides the methods mentioned above, there are many other works that
have been done in the field of dimensionality reduction. Supervised
Local Subspace Learning ($SL^2$) \cite{huang2011supervised} learns a
mixture of local tangent spaces that are robust to under-sampled
regions for continuous head pose estimation, so that it can avoid
overfitting and be robust to noise. Linear Spherical Discriminant
Analysis (LSDA) \cite{tang2012partially} performs discriminant
analysis based on the cosine distance metric to improve speaker
clustering performance. By building a sparse projection matrix for
dimension reduction, Double Shrinking Algorithm (DSA)
\cite{zhou2013double} compresses image data on both dimensionality
and cardinality to obtain better embedding or classification
performance. Least-Squares Dimension Reduction (LSDR)
\cite{suzuki2013sufficient} adopts a squared-loss variant of mutual
information as a dependency measure to perform sufficient
dimensionality reduction. Wang et al. proposed an exponential
framework for dimensionality reduction \cite{wang2014general}. By
using matrix exponential to measure data similarity, this framework
emphasizes small distance pairs, and can avoid the small sample size
problem. Although all of these methods have their own merits, none
of them solves the above mentioned two problems.

In this paper, we propose a novel supervised dimensionality
reduction method called Manifold Partition Discriminant
Analysis (MPDA), which solves the above two problems.
In MPDA, pairwise differences and piecewise
regional consistency are considered simultaneously,
so that the manifold structure can be well preserved.
MPDA aims to find a linear embedding space
where the within-class similarity is achieved along the
direction that is consistent with the local variation
of the data manifold, while nearby data belonging to
different classes are well separated.
Compared with existing methods, MPDA has several desirable
properties that should be highlighted:
\begin{itemize}
\item MPDA partitions the data manifold into a number of
non-overlapping linear subspaces and discovers regional
manifold structures in a piecewise manner.
\item With the partitioned manifold,
MPDA is able to construct tangent spaces with
varied numbers of dimensions. This provides MPDA with more flexibility
to handle non-uniformly distributed or complex data.
\item By using the first-order Taylor expansion,
MPDA establishes a manifold representation
which is characterized by both the pairwise differences
and piecewise regional consistency of the underlying manifold.
\item Thanks to the proposed manifold representation, MPDA
improves the measure of within-class similarity, and is able to
obtain a projection that is consistent with the local
variation of the underlying manifold.
\end{itemize}

The rest of this paper is organized as follows.
In Section \ref{sec:graph}, we briefly introduce the graph Laplacian based
framework, under which many supervised dimensionality reduction
methods can be considered within the same category. Then the
Manifold Partition Discriminant Analysis
(MPDA) algorithm is presented in Section \ref{sec:MPDA}.
Section \ref{sec:dis} discusses the connection
and difference between MPDA and related works.
In Section \ref{sec:exp}, MPDA is tested
on multiple real-world data sets compared with
existing supervised dimensionality reduction algorithms. Finally,
we give concluding remarks in Section \ref{sec:con}.

\section{Graph Laplacian Based Framework for Discriminant Analysis} \label{sec:graph}
Representing data on a specific graph is a popular
way to characterize the relationships among data points.
Given an undirected weighted $G = \{X, W\}$ with a vertex
set X and a symmetric weight matrix $W \in \mathbb{R}^{n \times n}$,
these relationships can be easily characterized by $G$,
where each example serves as a vertex
of $G$, and $W$ records the weight on the edge of each pair of vertices.
Generally, if two examples $\bm{x}_i$ and $\bm{x}_j$ are ``close", the
corresponding weight $W_{ij}$ is large, whereas if they
are ``far away", then the $W_{ij}$ is small. Provided a certain $W$,
the intrinsic geometry of graph $G$ can be represented by the Laplacian
matrix \cite{chung1997spectral}, which is defined as
\begin{equation} \label{LapMtrx}
L = D - W,
\end{equation}
where $D$ is a diagonal matrix with the $i$-th diagonal element
being $D_{ii} = \sum_{j\neq i}W_{ij}$. The Laplacian matrix
is capable of representing certain geometry of data according
to a specific weight matrix. This property is very helpful for
developing dimensionality reduction methods.

Let $X$ be a data set consisting of $n$ examples and labels,
$ \{ (\bm{x}_i,y_i) \}_{i=1}^n $, where $ \bm{x}_i \in \mathbb{R}^d$
denotes a \emph{d}-dimensional example, $y_i \in \{ 1,2,\ldots,C \}$ denotes the
class label corresponding to $\bm{x}_i$, and \emph{C} is the total
number of classes. Classical LDA aims to find an optimal
linear projection $\bm{t}$ along which the between-class scatter
is maximized and the within-class scatter is
minimized \cite{fukunaga1990introduction}.
The objective function of LDA can be written as:
\begin{equation} \label{obj LDA}
\bm{t}^* = \arg \max _{\bm{t}} \frac{ \bm{t} ^\top S_b \bm{t} }{ \bm{t} ^\top S_w \bm{t}},
\end{equation}
where $\top$ denotes the transpose of a matrix or a vector,
$S_b$ and $S_w$ denote the between-class and
within-class scatter matrices, respectively.
The definitions of $S_b$ and $S_w$ are given as follows:
\begin{equation} \label{Sb}
S_b = \sum_{c = 1} ^C n_c (\bm{\mu}_c - \bm{\mu}) (\bm{\mu}_c - \bm{\mu}) ^{\top},
\end{equation}
\begin{equation}
S_w = \sum_{c = 1} ^C \sum_{\{i | y_i = c \}} (\bm{x}_i - \bm{\mu}_c) (\bm{x}_i - \bm{\mu}_c) ^{\top},
\end{equation}
where $n_c$ is the number of data from the \emph{c}-th class,
$\bm{\mu} = \frac{1}{n} \sum _{i=1}^{n} \bm{x}_i$ is the mean
of all the data points, and $\bm{\mu}_c = \frac{1}{n_c}
\sum _{\{i | y_i = c \}} \bm{x}_i$ is the mean of the data from class \emph{c}.
Apart from the above formulations, $S_b$ and $S_w$ can also
be formulated via the graph Laplacian \cite{sugiyama2007dimensionality}:
\begin{align}
S_b=&\mathop{\sum}\limits_{ij}
W_{ij}^b \| \bm{x}_i-\bm{x}_j \|^2
=2X L^b X^\top, \label{LapSb} \\
S_w=&\mathop{\sum}\limits_{ij}
W_{ij}^w  \| \bm{x}_i-\bm{x}_j \|^2
=2X L^w X^\top, \label{LapSw}
\end{align}
where $L^w$ and $L^b$ are the Laplacian
matrices constructed by the weight matrices
$W^w$ and $W^b$ with
\begin{align}
W_{ij}^b =& \left\{ {\begin{array}{cl}
(1/n - 1/n_c) & if~y_i=y_j=c \\
1/n & if~y_i \neq y_j, \\
\end{array}} \right. \nonumber \\
W_{ij}^w =& \left\{ {\begin{array}{cl}
1/n_c  & if~y_i=y_j=c \\
0 & if~y_i \neq y_j.
\end{array}} \right. \nonumber
\end{align}
The objective function (\ref{obj LDA}) can be converted
to a generalized eigenvalue problem:
\begin{equation}\label{eigenProb}
X L^w X^\top\bm{t}=\lambda X L^b X^\top \bm{t}
\end{equation}
whose solution can be easily given by the eigenvector with
respect to the largest eigenvalue.
From the above formulations, it is clear that
the graph Laplacian plays a key role in deriving LDA, where
the weight matrices $W^b$ and $W^w$ measure
the similarity of each pair of data points,
and their characteristics varies as the criterion
of similarity changes. This provides
a general and flexible framework to develop
new dimensionality reduction algorithms by
constructing appropriate Laplacian matrices.

In order to improve the performance of LDA,
many local structure based extensions of LDA have been
proposed in the recent decades. Representative
methods include Marginal Fisher Analysis (MFA) \cite{yan2007graph},
Locality Sensitive Discriminant Analysis (LSDA) \cite{cai2007locality},
Local Fisher Discriminant Analysis (LFDA) \cite{sugiyama2007dimensionality}, etc.
Unlike traditional LDA, they compute the
between-class and within-class scatter based on local data structures
rather than the global mean values. Although these methods improve
the performance of discriminant analysis by
solving the problem caused by the improper assumption adopted by LDA,
none of them extends beyond the graph Laplacian based framework.
Their differences merely lie in the different ways of constructing
the Laplacian matrices $L^b$ and $L^w$.

In spite of its effectiveness, the graph Laplacian based
framework still has several limitations. The between-class
and within-class scatter are computed by only aggregating all pairwise
differences between data points across
the entire graph, whereas the regional consistency, which
is reflected by the regional structure around a local area
of the underlying manifold, is ignored.
Moreover, by minimizing the aggregation of
within-class data pairs (\ref{LapSw}),
the objective function (\ref{obj LDA}) tends to
find a direction along which some ``averaged"
within-class similarity is achieved.
However, it is unclear that how the ``averaged" similarity
can precisely reflect the topological and
geometrical structure of the underlying manifold.

\section{Manifold Partition Discriminant Analysis} \label{sec:MPDA}
In this section, we propose a novel supervised dimensionality reduction
algorithm named Manifold Partition Discriminant Analysis
(MPDA). Unlike previous methods that mainly rely on the
graph Laplacian \cite{yan2007graph,cai2007locality,sugiyama2007dimensionality},
MPDA exploits both pairwise differences and
piecewise regional consistency to preserve the manifold
structure. It aims to find a linear embedding space
where the within-class similarity is achieved along the
direction that is consistent with the local variation
of the data manifold, while nearby data belonging to
different classes are well separated. To this end,
we first need to extract the piecewise
consistency from the data manifold, which can be
achieved by partitioning the data manifold into
non-overlapping pieces, and estimating tangent spaces for each piece.
Then we can represent the data manifold by
combining pairwise differences with piecewise consistency.
The resultant manifold representation is able to
characterize the local variation of the data manifold,
and improve the measure of within-class similarity,
which eventually leads to the MPDA algorithm.
Specifically, we mainly solve the following problems:
\begin{itemize}
\item[\textit{P1}] How to partition the data manifold into
a number of non-overlapping pieces, and estimate
an accurate tangent space?
\item[\textit{P2}] How to combine pairwise differences
with piecewise regional consistency in representing the data manifold?
\item[\textit{P3}] How to find a linear subspace
where the within-class similarity is achieved along the
direction that is consistent with 
the local variation of the data manifold?
\end{itemize}
Next, we first solve \textit{P2} and \textit{P3} in Section
\ref{sec:maniRep} and \ref{sec:Alg}, respectively, and defer the treatment
of \textit{P1} to Section \ref{sec:Part}.

\subsection{Manifold Representation} \label{sec:maniRep}
In order to combine pairwise differences with piecewise regional
consistency in representing the data manifold, we are interested in
estimating a function $f$ defined on an \emph{m}-dimensional smooth
manifold $\mathcal{M}$, where $\mathcal{M}$ is embedded in
$\mathbb{R}^d$. This function $f$ can serve as a direct connection
between the data representation in $d$ and $m$-dimensional spaces.
For simplicity, we first consider to represent data in a
one-dimensional Euclidean space $\mathbb{R}$. Define $f:\mathbb{R}^d
\rightarrow \mathbb{R}$ as a function along the manifold
$\mathcal{M}$. Let $\mathcal{T}_{\bm{x}_0} \mathcal{M}$ be the
tangent space of $\bm{x}_0$ on $\mathcal{M}$, where ${\bm{x}_0} \in
\mathbb{R}^d$ is a single point on the manifold $\mathcal{M}$.
According to the first-order Taylor expansion at $\bm{x}_0$, $f$ can
be expressed as follows
\cite{simard2012transformation,lin2013parallel,sun2013tangent}:
\begin{equation*}
f(\bm{x}) = f(\bm{x}_0) + \bm{v}_{\bm{x}_0}^\top {\bm{u}_{\bm{x}_0}(\bm{x})}
+ O({\lVert \bm{x} - \bm{x}_0 \rVert}^2),
\end{equation*}
where $\bm{u}_{\bm{x}_0}(\bm{x}) = T_{\bm{x}_0}^\top (\bm{x} -
\bm{x}_0)$ is an \emph{m}-dimensional vector which gives a
representation of $\bm{x}$ in the tangent space
$\mathcal{T}_{\bm{x}_0} \mathcal{M}$. $T_{\bm{x}_0} \in
\mathbb{R}^{d \times m}$ is a matrix formed by the orthonormal bases
of $\mathcal{T}_{\bm{x}_0} \mathcal{M}$, and characterizes the
regional consistency of the manifold structure around $\bm{x}_0$.
Generally, $T_{\bm{x}_0}$ can be estimated by performing PCA on the
neighborhood of $\bm{x}_0$
\cite{min2004locality,Zhang02principalmanifolds}.
$\bm{v}_{\bm{x}_0}$ is an \emph{m}-dimensional tangent vector and
represents the manifold derivative of $f$ at $\bm{x}_0$ with respect
to $\bm{u}_{\bm{x}_0}(\bm{x})$, which reflects the local variation
of the manifold at $\bm{x}_0$.

Given two nearby data points $\bm{z}$ and $\bm{z}'$
lying on the manifold $\mathcal{M}$,
we can use the first-order Taylor expansion at $\bm{z}'$
to express $f(\bm{z})$ as follows:
\begin{equation} \label{TaylorExp}
f(\bm{z})=f(\bm{z}') + \bm{v}_{\bm{z'}}^\top T_{\bm{z'}}^\top (\bm{z} - \bm{z}')
+ O({\lVert \bm{z} - \bm{z}' \rVert}^2).
\end{equation}
If $\mathcal{M}$ is smooth enough,
the second-order derivatives of $f$ tend to vanish.
Furthermore, when $\bm{z}$ and $\bm{z}'$ are close to each other,
${\lVert \bm{z} - \bm{z}' \rVert}^2$
becomes very small. Therefore, the remainder in (\ref{TaylorExp})
can be omitted, which leads to:
\begin{equation} \label{maniExpr}
f(\bm{z}) \approx f(\bm{z}') +
\bm{v}_{\bm{z'}}^\top T_{\bm{z'}}^\top (\bm{z} - \bm{z}').
\end{equation}
With the above results, it is clear that for any nearby data points
$\bm{z}$ and $\bm{z}'$ lying on the manifold $\mathcal{M}$, the
low-dimensional embeddings $f(\bm{z})$ and $f(\bm{z}')$ should
satisfy (\ref{maniExpr}), and the difference between both sides of
(\ref{maniExpr}) should be as small as possible. This can serve as a
good criterion to preserve the manifold structure, which establishes
the connection between each pair of nearby data points.

\begin{figure}[t]
  \centering
  \includegraphics[width=250pt]{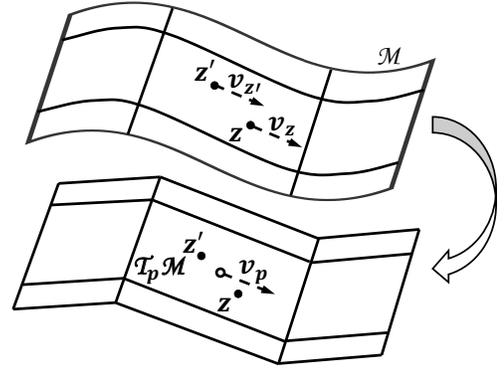}
  \caption{Conceptual illustration of the manifold partition strategy.
  \label{fig.MP}}
\end{figure}

Assume that the data manifold can be well approximated
by the union of a number of non-overlapping linear subspaces.
In this case, each linear subspace can serve as a tangent space,
and each tangent space has a tangent vector.
With the partitioned manifold, we are able to construct tangent
spaces and tangent vectors for each
linear subspace rather than each data point.
If $\bm{z}'$ lies in a tangent space $\mathcal{T}_{p} \mathcal{M}$
with a tangent vector $\bm{v}_{p}$, (\ref{maniExpr}) becomes:
\begin{equation} \label{maniExprPart}
f(\bm{z}) \approx f(\bm{z}') +
\bm{v}_{p}^\top T_{p}^\top (\bm{z} - \bm{z}'),
\end{equation}
where $T_{p}$ is estimated by performing PCA on the data falling
into $\mathcal{T}_{p} \mathcal{M}$. This can be justified by the
fact that the manifold derivative of a linear subspace is a constant
function. This means that for the data falling into the same linear
subspace, their corresponding tangent vectors are equal, and can be
represented by only one tangent vector $\bm{v}_p$. It is worth
noting that since PCA entails mean subtraction, each tangent space
estimated by PCA will have a separate mean. This seems to cause the
discrepancy of tangent spaces. However, this discrepancy is not a
problem in our case. Once the orthonormal basis $T_{p}$ has been
estimated, the effect of mean subtraction is just to center data to
the origin of the corresponding subspace. Notice that only the data
falling into $\mathcal{T}_{p} \mathcal{M}$ or those around
$\mathcal{T}_{p} \mathcal{M}$ are involved in the projection of
$T_p$. These data points implicitly reflect the mean of the
corresponding subspace. Therefore, we can directly use the
orthonormal basis $T_p$ to compute the projection without mean
subtraction. Figure \ref{fig.MP} illustrates the concept of the
above strategy (we call it the manifold partition). Intuitively,
after partitioning the manifold, $\mathcal{M}$ is approximated by
the union of the linear subspaces, where each linear subspace serves
as a tangent space. Therefore, (\ref{maniExprPart}) combines
pairwise differences with piecewise regional consistency in
representing the data manifold.

\subsection{The MPDA Algorithm} \label{sec:Alg}
Based on the above results, we propose our MPDA
algorithm. Consider a data set $X = \{ (\bm{x}_i,y_i) \}_{i=1}^n $
belonging to $C$ classes where $\bm{x}_i \in \mathbb{R}^d$
and $y_i \in \{ 1,2,\ldots,C \}$ is the class label associated with the
data point $\bm{x}_i$. Generally, we assume that data
in different classes are generated from different manifolds.
Provided that $X = \{ \bm{x}_1, \ldots, \bm{x}_n \}
= \bigcup_{p=1}^{P} X^p$ has been partitioned
into $P$ patches, where data of each patch
have the same class label, and we have obtained the
orthonormal basis matrices $\{T_p\}_{p=1}^P$ of tangent spaces
for each data patch. Our goal is to find an embedding
space where the within-class similarity is achieved along the
direction that is consistent with the local variation
of the data manifold, while nearby data belonging to
different classes are well separated.

In order to gather within-class data based on
the manifold structure, we first construct the
within-class graph $G=\{X,W\}$ to represent the
geometry of the data manifold. If $\bm{x}_i$ is among the $k$-nearest neighbors of
$\bm{x}_j$ with $y_i = y_j$, an edge connecting
$\bm{x}_i$ to $\bm{x}_j$ is added
with the weight $W_{ij}=W_{ji}=1$. If there is
no edge connecting $\bm{x}_i$ to $\bm{x}_j$, $W_{ij}=0$.
With the results in Section \ref{sec:maniRep}, for each
pair of nearby within-class data points,
we can obtain:
\begin{eqnarray}
f(\bm{x}_i) \approx f(\bm{x}_j) +
\bm{v}_{\pi_j}^\top T_{\pi_j}^\top (\bm{x}_i - \bm{x}_j), \label{TaylorZ1}\\
f(\bm{x}_j) \approx f(\bm{x}_i) +
\bm{v}_{\pi_i}^\top T_{\pi_i}^\top (\bm{x}_j - \bm{x}_i).  \label{TaylorZ2}
\end{eqnarray}
We require the difference between both sides of (\ref{TaylorZ1}) to
be as small as possible. In the scenario of linear dimensionality
reduction, $f(\bm{x})$ represents a one-dimensional embedding of
$\bm{x}$, and we aim to find a linear projection. To this end,
$f(\bm{x})$ is further approximated as a linear function
$f(\bm{x})=\bm{t}^\top \bm{x}$ where $\bm{t} \in \mathbb{R}^d$ is a
linear projection vector. Then, if nearby data points $\bm{x}_i,
\bm{x}_j$ belong to the same class, we can measure their similarity
as follows:
\begin{align} \label{regTerm1}
&\big( f(\bm{x}_i) - f(\bm{x}_j) - \bm{v}_{\pi_j}^\top T_{\pi_j}^\top
(\bm{x}_i - \bm{x}_j)\big)^2  \nonumber \\
=& \big( \bm{t}^\top (\bm{x}_i - \bm{x}_j) -
\bm{v}_{\pi_j}^\top T_{\pi_j}^\top (\bm{x}_i - \bm{x}_j)\big)^2,
\end{align}
where $\pi_i \in \{1,\ldots,P\}$ is an index
indicating the patch $\bm{x}_i$ belongs to.
Moreover, we also need to measure the similarity
between nearby tangent spaces.
By substituting (\ref{TaylorZ1}) into (\ref{TaylorZ2}),
we have:
\begin{equation*}
(T_{\pi_j} \bm{v}_{\pi_j} - T_{\pi_i}
\bm{v}_{\pi_i})^\top (\bm{x}_i - \bm{x}_j) \approx 0.
\end{equation*}
From the above equation, we know that the two vectors are
approximately perpendicular or the row vector $(T_{\pi_j}
\bm{v}_{\pi_j} - T_{\pi_i} \bm{v}_{\pi_i})^\top$ approximately
equals to a zero vector. However, the perpendicular case can not be
satisfied for every pair of nearby data points on the manifold. For
instance, consider there are three nearby data points on the
manifold. Each pair of them should satisfy the above equation, while
only two of them are, in general, justified in the perpendicular
case. On the other side, the case of zero row vectors can be
justified for all the data pairs, and leads to $T_{\pi_j}
\bm{v}_{\pi_j} \approx T_{\pi_i} \bm{v}_{\pi_i}$. Finally, by
multiplying both sides of this equation with $T_{\pi_i}^\top$ and
using $T_{\pi_i}^\top T_{\pi_i} = I$, it follows that:
\begin{equation} \label{reg2}
\bm{v}_{\pi_i} \approx T_{\pi_i}^\top T_{\pi_j} \bm{v}_{\pi_j}.
\end{equation}
It is clear that for each pair of nearby tangent spaces
the difference between both sides of (\ref{reg2})
should be as small as possible. Therefore, the similarity
between nearby tangent spaces can be measured as follows:
\begin{equation} \label{regTerm2}
\|\bm{v}_{\pi_i}-T_{\pi_i}^\top T_{\pi_j} \bm{v}_{\pi_j}\|_2^2.
\end{equation}

With the above results, the data manifold with respect to
each class can be estimated by relating data
with a discrete weight $W_{ij}$, which leads to an
objective function as follows:
\begin{eqnarray}
\begin{split}
\min_{\bm{t},\bm{v}} \mathop{\sum}\limits_{i,j}^{n} W_{ij}&
\Big[ \big( \bm{t}^\top (\bm{x}_i - \bm{x}_j) -
\bm{v}_{\pi_j}^\top T_{\pi_j}^\top (\bm{x}_i - \bm{x}_j)\big)^2 \\
&+ \gamma\|\bm{v}_{\pi_i}-T_{\pi_i}^\top T_{\pi_j} \bm{v}_{\pi_j}\|_2^2 \Big],
\end{split} \label{withinPatchObj}
\end{eqnarray}
where $\gamma$ is a trade-off parameter controlling the influence
between (\ref{regTerm1}) and (\ref{regTerm2}).
It is clear that if $\bm{x}_i$ and $\bm{x}_j$ belong to the
same class and fall into the same
tangent space, their similarity only depends on
their pairwise difference. If $\bm{x}_i$ and $\bm{x}_j$
belong to the same class but lie in different
tangent spaces, apart from the pairwise difference, their similarity
also depends on the angle between
$\bm{v}_{\pi_j}$ and $T_{\pi_j}^\top (\bm{x}_i - \bm{x}_j)$,
which means that $\bm{x}_i$ and $\bm{x}_j$ can be viewed as
similar data points when $\bm{v}_{\pi_j}$ and
$T_{\pi_j}^\top (\bm{x}_i - \bm{x}_j)$
have similar directions.
Since $\bm{v}_{\pi_j}$ reflects the varying direction of
the data manifold around $\bm{x}_j$, by optimizing (\ref{withinPatchObj}),
we can deem that the within-class similarity is achieved along the
direction that is consistent with the local variation
of the data manifold.



It is worth noting that the above derivation is based on
the first-order Taylor expansion of the function $f$.
If we employ the zero-order Taylor expansion,
the terms related to $\bm{v}_{\pi_i}~(i=1,\ldots,n)$ vanish.
Then the objective function is simplified as follows:
$$\min_{\bm{t}} \mathop{\sum}\limits_{i,j}^{n} W_{ij}
\big( \bm{t}^\top \bm{x}_i - \bm{t}^\top \bm{x}_j\big)^2=
\min_{\bm{t}} 2 \bm{t}^\top X L X^\top \bm{t},$$ where $L=D-W$ is
the Laplacian matrix and $D$ is a diagonal matrix with the $i$-th
diagonal element being $D_{ii} = \sum_{j\neq i}W_{ij}$. This
formulation is identical to the graph Laplacian based within-class
scatter (\ref{LapSw}). From the aspect of manifold approximations,
this means that in theory the proposed method is able to approximate
the underlying manifold with a smaller approximation error
$O(||\bm{x}_i-\bm{x}_j||^2)$ than the graph Laplacian whose
approximation error is $O(||\bm{x}_i-\bm{x}_j||)$. Compared with
(\ref{withinPatchObj}), the graph Laplacian based scatter fails to
consider the regional consistency that is explicitly parameterized
by the proposed manifold representation. Although it can implicitly
reflect regional relationships by minimizing the distance between
each pair of nearby data points, the graph Laplacian has no ability
to capture the regional consistency which is determined by all the
nearby data around a given data point. On the other hand, the
proposed manifold representation is capable of preserving both the
pairwise geometry and the piece-wise regional consistency, and thus
can capture more structural information from the data manifold than
the graph Laplacian. In other word, (\ref{withinPatchObj}) better
measures the within-class similarity than the graph Laplacian based
scatter (\ref{LapSw}), because it can extract the regional
consistency of each tangent space, and explicitly establishes the
connections among tangent spaces by estimating tangent vectors $\{
\bm{v}_p \}_{p=1}^P$.

Notice that although we assume that the data manifold can be
approximated by a union of piece-wise subspaces, it does not mean
that the proposed manifold representation is inferior to the graph
Laplacian. To verify this, we can split the within-class objective
function (\ref{withinPatchObj}) into two parts. The first part
includes the terms related to $\bm{v}$, and the second part has the
other terms. The piece-wise manifold assumption only affects the
first part, while the second part is still based on the generic
manifold assumption. In fact, the second part of
(\ref{withinPatchObj}) is just identical to the graph Laplacian
based within-class scatter. This implies that the proposed manifold
representation is at least as good as, if not better than, the graph
Laplacian, as each tangent vector $\bm{v}_{\pi_i}$ can be a zero
vector.

To separate data in different classes, we construct a between-class graph $G'=\{X,W'\}$.
If $y_i \neq y_j$, we add an edge between $\bm{x}_i$ and $\bm{x}_j$
with the weight $W'_{ij}=1/n$. If $y_i=y_j$,
the corresponding weight is set to be $W'_{ij}=A_{ij}(1/n - 1/n_c)$.
$n_c$ is the number of data points from
the \emph{c}-th class, and $A_{ij}$ is a weight that
indicates the similarity between $\bm{x}_i$ and $\bm{x}_j$,
whose definition is given as follows:
\begin{equation*}
A_{ij} = \left\{ {\begin{array}{cc}
\exp(- \frac{{\| \bm{x}_i - \bm{x}_j \|}^2}{\sigma_i \sigma_j}) &
~~\mbox{if}~i \in \mathcal{N}_{k}(j)~\mbox{or}~j \in \mathcal{N}_{k}(i) \\
0 & ~~\mbox{else},
\end{array}} \right.
\end{equation*}
where $\mathcal{N}_{k}(i)$ denotes the $k$-nearest
neighbor set of $\bm{x}_i$, and $\sigma_i$ is heuristically set to
be the distance between $\bm{x}_i$ and its $k$-th nearest neighbor.
Then we can formulate the following objective function
to separate nearby between-class data points:
\begin{align}
\max_{\bm{t}} \mathop{\sum}\limits_{i,j}^{n} W_{ij}'
(\bm{t}^\top \bm{x}_i - \bm{t}^\top \bm{x}_j)^2 \label{betweenObj}.
\end{align}
The methods for constructing $G'$ have been well studied in
the literature \cite{yan2007graph,cai2007locality}. Here, we
employed the one in \cite{sugiyama2007dimensionality} because
of its effectiveness in enhancing the between-class separability.

It is easy to see that (\ref{withinPatchObj}) can be
reformulated as a canonical matrix quadratic as
$ (\begin{array}{*{20}{c}}
{{\bm{t}^ \top }}&{{\bm{v}^ \top }}
\end{array})S{(\begin{array}{*{20}{c}}
{{\bm{t}^ \top }}&{{\bm{v}^ \top }}
\end{array})^ \top } $
where $S$ is a $(d + mP) \times (d + mP)$ positive semi-definite
matrix and $\bm{v} = (\bm{v}_{1}^\top,\bm{v}_{2}^\top,
\ldots,\bm{v}_{P}^\top)^\top$. Due to the space limitation, the
detailed derivation of $S$ is provided in the supplementary
material, which is a modification of the derivation of a similar
quantity used in \cite{sun2013tangent}. By simple algebra
formulations, (\ref{betweenObj}) can also be reduced to ${\left(
{\begin{array}{*{20}{c}} \bm{t} \\ \bm{v} \end{array}} \right)^ \top
} \left( {\begin{array}{*{20}{c}} {2 X L' X^\top}& \bm{0} \\ \bm{0}
& \bm{0}
\end{array}} \right)
\left( {\begin{array}{*{20}{c}} \bm{t} \\ \bm{v} \end{array}} \right)
=(\begin{array}{*{20}{c}}
{{\bm{t}^ \top }}&{{\bm{v}^ \top }}
\end{array})S'{(\begin{array}{*{20}{c}}
{{\bm{t}^ \top }}&{{\bm{v}^ \top }}
\end{array})^ \top },$
where $L'$ is the Laplacian matrix constructed by $W'$.
In order to preserve the manifold structure while
separating nearby between-class data points, we can
optimize the objective functions (\ref{withinPatchObj})
and (\ref{betweenObj}) simultaneously, which leads to
the following objective function:
\begin{equation} \label{TSD_Obj}
\arg \max_{\bm{f}} \frac{\bm{f}^\top S' \bm{f}}
{\bm{f}^\top (S + \alpha I) \bm{f}},
\end{equation}
where we have defined $\bm{f} = (\bm{t}^\top, \bm{v}^\top)^\top$, and
the Tikhonov regularizer with a trade-off parameter $\alpha$
has been employed to avoid the numerical singularity of $S$.

The optimization of (\ref{TSD_Obj}) is
achieved by solving a generalized eigenvalue problem:
\begin{equation} \label{EigenDecomp}
S' \bm{f} = \lambda (S + \alpha I) \bm{f}
\end{equation}
whose solution is the eigenvector
$\bm{f}^* = (\bm{t}^{*\top}, \bm{v}^{*\top})^\top$
with respect to the largest eigenvalue.
Then we can use the first part of $\bm{f}^*$ to obtain a one-dimensional
embedding of any $\bm{x} \in \mathbb{R}^d$ by computing $b = \bm{t}^{*\top} \bm{x}$.
If we want to project $\bm{x}$
into an $m$-dimensional subspace, we can just compute
$m$ eigenvectors $\bm{f}_1,\ldots,\bm{f}_m$ corresponding
to the $m$ largest eigenvalues of (\ref{EigenDecomp}). Then the $m$-dimensional
embedding $\bm{b}$ of $\bm{x}$ is computed as
$\bm{b} = T^\top \bm{x}$, where $T = (\bm{t}_1,\ldots,\bm{t}_m).$
Algorithm \ref{pseudo-code} gives the pseudo-code for MPDA.

\begin{algorithm}[tb]  
   \caption{MPDA}
   \label{pseudo-code}
\begin{algorithmic}
   \STATE {\bfseries Input:}
   \STATE ~~~~Labeled data $ \{ \bm{x}_i | \bm{x}_i \in \mathbb{R}^d \},$
   \STATE ~~~~Class labels $\{y_i | y_i \in \{ 1,2,\ldots,C \} \}_{i=1}^n $;
   \STATE ~~~~Dimensionality of embedding space $m$ ($1 \leq m \leq d$);
   \STATE ~~~~Trade-off parameters $\gamma$ ($\gamma > 0$).
   \STATE {\bfseries Output:}
   \STATE ~~~~$d \times r$ transformation matrix $T$.
   \STATE
   \STATE Apply certain method to partition the
   data in each class into a total of $P$ patches $\{X^p\}_{p=1}^P$;
   \FOR{$p=1$ {\bfseries to} $P$}
   \STATE Construct $T_p$ by performing PCA on $X^p$;
   \ENDFOR
   \STATE Construct the within-class graph $G$ and
   the between-class graph $G'$;
   \STATE Compute the eigenvectors $\bm{f}_1,\bm{f}_2,\ldots,\bm{f}_m$
   of (\ref{EigenDecomp}) with respect to the top $m$ eigenvalues;
   \STATE $T = (\bm{t}_1,\bm{t}_2,\ldots,\bm{t}_m)$.
\end{algorithmic}
\end{algorithm}

\subsection{Partitioning the Manifold} \label{sec:Part}
In this section, we propose a manifold partition algorithm
to solve the last problem (\textit{P1}).
Since tangent spaces are linear subspaces in essence,
the better the data manifold can be linearly
approximated by the partitioned pieces,
the more accurately the resultant tangent spaces can
reflect the regional consistency of the underlying manifold.
In order to estimate tangent spaces which
approximately lie on the manifold surface,
we first need to introduce a criterion to measure
the linearity of subspaces.

Given a data set $X$ as well as its
pairwise Euclidean distance matrix $D^E$ and geodesic distance
matrix $D^G$ (approximated by the shortest path algorithms such as
Dijkstra's algorithm), we can measure the degree of
linearity between two data points $\bm{x}_i$ and $\bm{x}_j$
by computing the ratio $R_{ij} = D^G_{ij}/D^E_{ij}$,
which is also referred to as the \textit{tortuosity} \cite{clennel97}.
$D^E_{ij}$ is the Euclidean distance between $\bm{x}_i$ and $\bm{x}_j$,
and $D^G_{ij}$ is their geodesic distance.
It is clear that $D^G_{ij}$ is never smaller than $D^E_{ij}$.
If $D^G_{ij} \approx D^E_{ij}$, then
$R_{ij} \approx 1$ and we can deem that
$\bm{x}_i$ and $\bm{x}_j$ lie on a straight line.
When it comes to a data patch $X^p$, we can measure its linearity
as follows:
\begin{equation}
R^p = \frac{1}{N^2_p} \sum\limits_{\bm{x}_i \in X^p}
\sum\limits_{\bm{x}_j \in X^p} R_{ij}, \label{linearScore}
\end{equation}
where ${N_p}$ denotes the number of data in $X^p$. It is clear
that the smaller $R^p$ is, the better the data in
$X^p$ fit a linear subspace.

With the above measure of linearity, we can partition the manifold
by hierarchical clustering \cite{kaufman2009finding}. There are
mainly two branches of hierarchical clustering depending on their
search strategies. In this paper, we use the top-down hierarchical
divisive clustering rather than the bottom-up hierarchical
agglomerative clustering because of two reasons. For one thing, if
we need to partition the data set into $P$ patches, as $P$ is
usually much smaller than the number of data points, top-down
methods are more efficient than bottom-up ones. For another,
top-down methods tend to construct patches with the same or similar
sizes. As a result, the tangent spaces estimated by these patches
tend to have similar dimensionalities, which fits the manifold
assumption better. Specifically, given a data set $X =
\{\bm{x}_1,\ldots,\bm{x}_{N} \}$, our top-down partition algorithm
aims to partition $X$ into a number of patches (subsets) until there
is no patch (subset) containing more than $M$ data points, which
consists of the following steps:
\begin{enumerate}
\item
Initialize $P=1$, $X = \{X^p\}_{p=1}^P = \{X^1\} =
\{\bm{x}_1,\ldots,\bm{x}_{N_1} \} $, where ${N_1}=N$. Compute the
Euclidean distance matrix $D^E$, the geodesic distance matrix $D^G$
(approximated by the shortest path algorithms such as Dijkstra's
algorithm), and the patch linearity $R^1$ according to
(\ref{linearScore}).
\item  From $\{X^p\}_{p=1}^P$, select the patch $X^p~(p \in {1,\ldots,P})$ having the highest
value of $R^p \cdot N_p$. From $X^p$, select two data points
$\bm{x}_l$ and $\bm{x}_r$ having the largest geodesic distance
$D^G_{lr}$. Create two new patches $X^p_l=\{ \bm{x}_l \}$ and $X^p_r
= \{ \bm{x}_r \}$. Update $X^p\leftarrow X^p \setminus \{\bm{x}_l,
\bm{x}_r\}$.
\item Construct the $k'$-nearest neighbor sets of $X^p_l$
and $X^p_r$ denoted by $\mathcal{N}^p_l$ and $\mathcal{N}^p_r$, respectively.
Construct the joint neighbor set $\mathcal{N}^p_{joint} = \mathcal{N}^p_l \cap \mathcal{N}^p_r$.
Update $\mathcal{N}^p_l \leftarrow \mathcal{N}^p_l \setminus \mathcal{N}^p_{joint}$,
$\mathcal{N}^p_r \leftarrow \mathcal{N}^p_r \setminus \mathcal{N}^p_{joint}$.
\item Update $X^p_l\leftarrow X^p_l \cup (\mathcal{N}^p_l \cap X^p)$,
$X^p\leftarrow X^p \setminus \ (\mathcal{N}^p_l \cap X^p)$,
$X^p_r\leftarrow X^p_r \cup (\mathcal{N}^p_r \cap X^p)$,
$X^p\leftarrow X^p \setminus \ (\mathcal{N}^p_r \cap X^p)$.
\item Compute the patch linearity $R^p_l$ and
$R^p_r$ for $X^p_l$ and $X^p_r$, respectively.
Let $N_l$ and $N_r$ be the number of data in $X^p_l$
and $X^p_r$. If $R^p_l \cdot N_l > R^p_r \cdot N_r$, update
$X^p_r \leftarrow X^p_r \cup \mathcal{N}^p_{joint}$,
or update $X^p_l \leftarrow X^p_l \cup \mathcal{N}^p_{joint}$ otherwise.
Repeat steps $3) \sim 5)$ until $X^p = \bm{\varnothing}$.
\item $X^p$ has been partitioned into $X^p_l$ and $X^p_r$.
 Update $P \leftarrow P+1$, $X^p\leftarrow X^p_l$,
$X^P\leftarrow X^p_r$. Go to step 2), until there is no patch having
$N_p > M$, where $M$ is the maximum patch size.
\end{enumerate}

Generally, in order to obtain the patch in which data
lie in a linear subspace, we should divide the patch with the largest
$R^p$ in each turn of partition. In our algorithm, we
combine the patch linearity $R^p$ and its size $N_p$
together to select the patch that should be
further divided, because the scope of subspaces should be small
enough so that the Taylor expansion in (\ref{TaylorZ1})
and (\ref{TaylorZ2}) can be justified.
Two parameters in the proposed partition algorithm
should be determined, i.e.,
the neighborhood size $k'$ and the maximum patch
size $M$. It is worth noting that to estimate tangent spaces accurately, each patch
should satisfy two competing requirements. On the one
hand, we should keep sufficient data in each patch
so that the tangent space can be well estimated. On
the other hand, the patch should be small enough to preserve the local
manifold structure. Therefore, we use $M$ rather
than the number of subspaces $P$ as the threshold
to control the termination of the algorithm.

\begin{figure}[t!] 
\centering
\includegraphics[width=200pt]{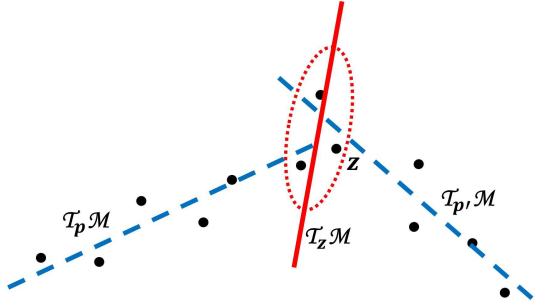}
\caption{An example of how performing PCA on the
  fixed-sized neighborhood fails to capture the correct tangent space.
  Dashed lines indicate the tangent spaces of two patches
  $X^p$ and $X_{p'}$.
  The dotted ellipse indicates the two-nearest neighborhood of $\bm{z}$.
  The solid line shows the tangent space estimated by
  performing PCA on the two-nearest neighborhood of $\bm{z}$.}
\label{Fig.kNNFail}
\end{figure}

Besides extracting the piecewise regional consistency,
partitioning the manifold can provide
additional benefits. It is clear that
an accurate estimation of tangent spaces is crucial
for tangent space based methods. Usually,
tangent spaces are estimated by performing PCA on
the $k$-nearest neighbors of each data point.
This approach fixes the neighborhood size, which
may fail to estimate the correct tangent spaces when data are
sampled non-uniformly or the manifold has a varying
curvature. In contrast, the proposed MPDA method is more
likely to get a robust estimation,
because PCA is performed on the data in each linear
subspace where data naturally lie on the
manifold surface. Figure \ref{Fig.kNNFail} shows
an example that performing PCA on the fixed-sized neighborhood
fails to capture the correct tangent space. As can be seen,
$\mathcal{T}_p \mathcal{M}$ and $\mathcal{T}_{p'} \mathcal{M}$
reflect the correct manifold structure,
whereas $\mathcal{T}_{\bm{z}} \mathcal{M}$ computed
by $\bm{z}$ and its two-nearest neighbors is incorrect.
In addition, real data are often complex whose underlying
manifold dimensionality could vary at different regions.
Therefore, it would be better to adjust the manifold
dimensionality for different parts of the manifold instead
of setting a fixed one. As the number of data varies in
each linear subspace, MPDA adaptively determines the number of
dimensions of each linear subspace by simply employing PCA to preserve
certain percentages of energy, say 95\%.
This provides MPDA with more flexibility
to handle complex data in practice.

\subsection{Pairwise-variate MPDA}
For now, we have presented the MPDA algorithm which considers both
preserving the manifold structure and distinguishing data from
different classes. Since the data manifold is partitioned into a
number of non-overlapping tangent spaces, MPDA discovers the
regional consistency of the data manifold in a piecewise manner,
where the manifold partition strategy plays a key role in deriving
MPDA. If we relax the piece-wise manifold assumption to the general
one, and directly derive the proposed method from (\ref{maniExpr})
rather than (\ref{maniExprPart}) without partitioning the manifold,
we can obtain a Pairwise-variate MPDA (PMPDA). Then, the objective
function (\ref{withinPatchObj}) becomes:
\begin{align}
\min_{\bm{t},\bm{v}} \mathop{\sum}\limits_{i,j}^{n} W_{ij}&
\Big[ \big( \bm{t}^\top (\bm{x}_i - \bm{x}_j) -
\bm{v}_{\bm{x}_j}^\top T_{\bm{x}_j}^\top (\bm{x}_i - \bm{x}_j)\big)^2 \nonumber\\
&+ \gamma\|\bm{v}_{\bm{x}_i}-T_{\bm{x}_i}^\top T_{\bm{x}_j} \bm{v}_{\bm{x}_j}\|_2^2 \Big],
\label{withinObj}
\end{align}
where the orthonormal basis matrix $T_{\bm{x}_i}$ is computed by performing
PCA on the $k$-nearest within-class neighbors of $\bm{x}_i$.

Similar to (\ref{withinPatchObj}), (\ref{withinObj})
can also be reformulated as a quadratic form
$(\begin{array}{*{20}{c}}
{{\bm{t}^ \top }}&{{\bm{v}^ \top }}
\end{array})S_p{(\begin{array}{*{20}{c}}
{{\bm{t}^ \top }}&{{\bm{v}^ \top }}
\end{array})^ \top } $
where $S_p$ is a $(d + mn) \times (d + mn)$ positive
semi-definite matrix, and $\bm{v} = (\bm{v}_{\bm{x}_1}^\top,\bm{v}_{\bm{x}_2}^\top,
\ldots,\bm{v}_{\bm{x}_n}^\top)^\top$.
The rest steps of PMPDA are just the same as MPDA except
that $S$ is replaced by $S_p$. Finally, PMPDA solves
the following generalized eigenvalue problem:
\begin{equation}
S' \bm{f} = \lambda (S_p + \alpha I) \bm{f}. \label{EigenPMPDR}
\end{equation}
Compared with MPDA, PMPDA no longer needs to partition the manifold,
but has to estimate tangent vectors and tangent spaces for each data
point, while MPDA estimates only $P$ of them. On the one hand, PMPDA
is more effective to preserve the manifold geometry, because it is
based on a more general manifold assumption.
On the other hand, PMPDA has to determine tangent
vectors and construct tangent spaces at each data point,
which not only leads to tremendous computational overheads, but
results in severe storage problems when it is performed on large
data sets. Consequently, PMPDA can only be performed on small data
sets and is hardly practical. In brief, PMPDA is able to make a
better manifold estimation at the expense of its efficiency, and the
strategy of partitioning the manifold can be viewed as a trade-off
between effectiveness and efficiency of the manifold estimation.

Like PMPDA, some tangent space based methods such as
TSIMR \cite{sun2013tangent} and PFE \cite{lin2013parallel}
also suffer from similar computational
and storage problems. This implies that
although the manifold partition strategy is a crucial part of MPDA,
we can also apply it to the tangent space based methods
to make them more efficient. We provide some
preliminary results in the supplementary material.

\subsection{Time Complexity}
In this section, we briefly analyze the computational complexity
of both PMPDA and MPDA. The main computational costs of PMPDA
lie in building tangent spaces for $n$ data points and solving
the generalized eigenvalue problem.
PMPDA takes $O((d^2k+k^2d)\times n)$ for estimating
$n$ tangent spaces by performing PCA on the $k$-nearest
neighborhood of each data point.
Note that we can obtain at most $k$ meaningful
orthonormal bases for each tangent space,
since there are only $k+1$ data points as the inputs of PCA.
Therefore, the dimensionalities of tangent vectors and
tangent spaces are at most $k$. This means that
PMPDA takes $O((d+kn)^3)$ for solving the
generalized eigenvalue problem (\ref{EigenPMPDR}).

MPDA first partitions the manifold
into $P$ linear subspaces, whose time complexity is dominated by computing
the geodesic distance matrix $D^G$ and the hierarchical divisive
clustering for data in each class. Computing $D^G$ based on a $k'$-NN graph
by the Dijkstra's algorithm with Fibonacci heaps takes $O(n^2 \log n + k' n^2/2)$.
The computational complexity
of the hierarchical divisive clustering can be approximated as
$O(\sum_{c=1}^C \sum_{p=1}^{P_c} (2^p(n_c/2^p))^2) \approx O(\sum_{c=1}^C n_c^2)$,
where $n_c$ is the number of data in
the \emph{c}-th class, and $P_c$ is the number of patches
partitioned from the data in the $c$-th class with $\sum_{c=1}^C P_c = P$.
Then MPDA takes $O(\sum_{p=1}^P (d^2N_p+{N_p}^2d))$
for estimating $P$ tangent spaces
and $O((d+\sum_{p=1}^P N_p)^3)=O((d+n)^3)$ for
solving the generalized eigenvalue problem.

With the above results, we can find that the
most consuming parts of PMPDA and MPDA
lie in the generalized eigenvalue decomposition.
PMPDA needs to decompose $S_p$,
a large matrix sized $(d + mn) \times (d + mn)$,
which will takes $O((d + kn)^3)$.
Compared with PMPDA, MPDA manipulates a much smaller matrix
$S$ sized $(d + mP) \times (d + mP)$ and only needs to estimate $P$ tangent
spaces rather than $n$. Since $P \ll n$, this leads to significant
computational savings.

\subsection{Further Improvement}
Based on the above analysis, it is clear that MPDA
avoids both computing tangent spaces for every
data point and solving the eigenvalue problem with
a large matrix, so that it have less computational
complexity. In fact, we can make it more scalable. Note that
we use the product $R^p \cdot N_p$
to determine the patch that should be further divided and
the manner that how this patch is divided, where $N_p$ is predominant to
control our partition algorithm.
When the number of data is very large, we can deem that the Euclidean
distance is approximately equal to the geodesic distance
within a small region, which leads to $R^p \approx 1$.
Therefore, the partition algorithm can be simplified by omitting the computation
of $D^G$ and $R^p$, so that we can save the time for performing Dijkstra's algorithm
and computing $R^p$.

In addition, the computational costs of MPDA can be
reduced by estimating tangent vectors and
tangent spaces only at anchor points. In this case, we
are interested in selecting a portion of points from the original data set
as the anchor points, and the rest can be
represented according to the first-order Taylor expansion at
their nearest anchor points. Therefore, the data
manifold can be estimated by using the anchor points only.
It is natural to
specify the center of each linear subspaces, which is not
necessary a data point among the training set, as the anchor
point. As a result, MPDA can be performed on only $P$
anchor points rather than the whole data set, such that
the corresponding computational complexity for solving
the generalized eigenvalue problem can be reduced to $O((d+P)^3)$.

Moreover, the two-stage strategy \cite{ye2005two}
can be adopted to further reduce the computational costs.
We can separate the generalized eigenvalue
problem (\ref{EigenDecomp}) into two stages. The first
stage maximizes (\ref{betweenObj}) via QR decomposition
to find its solution space.
The second one solves (\ref{EigenDecomp}) in the solution
space of (\ref{betweenObj}). Since $S'$ is just the extension
of $2X L' X^\top$, the rank of $S'$ is at most $d$.
Consequently, the time for solving (\ref{EigenDecomp})
can be reduced to $O(d^3)$. Please refer to \cite{ye2005two} for more details.

\section{Discussion} \label{sec:dis}
Several works have been done to manipulate
data in local subspaces for dimensionality reduction
\cite{teh2002automatic,verbeek2006learning,yang2008alignment,wang2011maximal}.
Basically, they share the same spirit in aligning
local subspaces to build a global coordinate,
where the connections of local subspaces
are considered implicitly. The main difference
between MPDA and these methods is that MPDA
constructs tangent spaces in a piecewise manner
and explicitly characterizes their connections
by estimating tangent vectors.

Local Linear Coordination (LLC) \cite{teh2002automatic}
and Coordinated Factor Analysis (CFA) \cite{verbeek2006learning}
construct linear subspaces through the mixture
of factor analyzers (MFA) which can serve
as an alternative way to partition the data manifold.
However, MFA is optimized by the expectation-maximization (EM)
algorithm, which can be slow and unstable. Moreover,
the number of factor analyzers and
the dimensionality of each linear subspace should be
specified as a priori knowledge, which are difficult to determine.
In contrast, the proposed manifold partition algorithm
for constructing linear subspaces is more efficient, and the dimensionality of
each linear subspace can be determined automatically
by using PCA.

Compared with MPDA, Maximal Linear Embedding
(MLE) \cite{wang2011maximal} also constructs
a number of linear subspaces based on the measure of linearity
but follows a different principle.
MLE prefers to construct the linear subspaces whose
sizes should be as large as possible, while MPDA constructs
a number of linear subspaces with similar and relatively
small sizes to justify the manifold assumption as well as the
Taylor expansion. Another difference between MPDA and
MLE is that although MPDA partitions the data manifold
to extract piecewise regional consistency,
it still use all the data to
discover the underlying manifold, whereas
MLE only uses a portion of data to obtain the resultant global coordinate.
This means that MPDA utilizes more information
from data sets than MLE.

Locally Multidimensional Scaling (LMDS) \cite{yang2008alignment} can
be seen as a sparsified version of LTSA. It constructs
tangent spaces based on a set of overlapping local subspaces
where the number of subspaces should be as small as possible.
This strategy allows LMDS to avoid estimating
tangent spaces for each data point, and thus makes LMDS more efficient
than LTSA. However, if the local subspaces are non-overlapping, LMDS
cannot work normally any more, because as an
alignment based method, it needs the overlapping parts
of local subspaces to serve as the implicit connections
for aligning a global coordinate.
In contrast, since MPDA explicitly characterizes
the connections among tangent spaces by estimating tangent vectors,
it can construct a global coordinate based on
non-overlapping local subspaces.

\begin{table}[!t]
\renewcommand{\arraystretch}{1.3}
\caption{Statistics of the data sets: $n$ is the number of data points,
$d$ is the data dimensionality, $C$ is the number of classes, and
$\delta$ is the percentages of training data.}
\label{table:config}     
\centering
\small
\begin{tabular}{c|cccc}
\hline\noalign{\smallskip}
Data Set & \ $n$ & \ $d$ & $C$ & $\delta$ \\
\noalign{\smallskip}\hline\noalign{\smallskip}
COIL20 & 1440 & 1024 & 20 & 25\%\\
COIL100 & 7200 & 1024 & 100 & 25\%\\
FaceDetection & 2000 & 361 & 2 & 25\% \\
MNIST & 4000 & 784 & 10 & 25\% \\
OptDigits & 5620 & 64 & 10 & 25\%\\
Semeion & 1593 & 256 & 10 & 25\%\\
Vehicle & 846 & 18 & 4 & 50\%\\
\noalign{\smallskip}\hline
\end{tabular}
\end{table}

\begin{table*}
\centering
\small
\caption{Average error rates (dimensionality) on different data sets.}
\label{table:performance}      
\begin{tabular}{l|ccccccc}
\hline\noalign{\smallskip}
Methods & COIL20 & COIL100 & FaceDetection & MNIST & OptDigits & Semeion & Vehicle \\
\noalign{\smallskip}\hline\noalign{\smallskip}
Baseline & 4.71\%(1024) & 10.49\%(1024) & 7.97\%(361) & 12.78\%(784)
         & 2.11\%(64)& 14.51\%(256) & 36.75\%(18) \\
PCA      & 3.24\%(23.95) & 7.48\%(38.55) & 4.85\%(23.45) & 10.90\%(32.75)
         & 2.03\%(35.05)& 11.83\%(34.75)    &36.62\%(13.95) \\
LDA      & 3.11\%(19) & 13.99\%(99) & 7.44\%(1) & 18.98\%(9)
         & 4.66\%(9) & 14.66\%(9)   & 26.67\%(3) \\
MFA      & 2.30\%(15.8) & 7.50\%(27.35) & 2.93\%(22.35) & 12.21\%(47.4)
         & 2.52\%(34.65)& 12.69\%(30.65)    & 20.20\%(11.15) \\
LSDA     & 3.31\%(19.6) & 8.79\%(27.65) & 3.54\%(27.65) & 11.77\%(34.65)
         & 2.52\%(28.45)& 12.66\%(19.1) & 22.09\%(10.6) \\
LFDA     & 1.89\%(17.2) & 7.39\%(33.1) & 2.82\%(44.8) & 13.53\%(34.75)
         & 2.33\%(27.95)& 12.22\%(29.2) & \bf{19.63\%(10.65)} \\
LLTSA    & 6.49\%(23.65) & 15.72\%(49) & 4.85\%(26.55) & 17.27\%(32)
         & 3.94\%(22.55)& 19.92\%(17.3) & 23.84\%(17.45) \\
LSDR     & 4.12\%(48.15) & - & 7.68\%(85.6) & 12.40\%(123.4)
& - & 14.30\%(120.05) & 36.37\%(14.45) \\
LPFE     & 3.23\%(50.05) & 9.60\%(155.35) & 5.74\%(71.85) & 18.42\%(124.3)
         & 8.68\%(28.1)& 21.01\%(112.6) & 49.43\%(11.4) \\
PMPDA    & \bf{1.45\%(14.45)} & -       & \bf{1.64\%(25.1)} & \bf{9.70\%(22)}
         & - & \bf{9.26\%(22.6)} & 22.27\%(11.6) \\
MPDA     & \bf{1.25\%(13.85)} & \bf{6.69\%(25.75)} & 2.15\%(26.9) & 10.09\%(28.7)
         & \bf{1.90\%(23.9)} & \bf{8.86\%(22.45)}   & \bf{19.55\%(8.9)} \\
\noalign{\smallskip}\hline
\end{tabular}\vspace*{-0.2in}
\end{table*}


\section{Experiment} \label{sec:exp}
\subsection{Real-World Data Sets} \label{sec:ExpReal}
We focus on supervised dimensionality reduction tasks and test the
proposed PMPDA and MPDA on multiple real-world data sets.
Comparisons are made with: 1) Classical baseline methods including
PCA and LDA; 2) Graph Laplacian based methods including Marginal
Fisher Analysis (MFA) \cite{yan2007graph}, Locality Sensitive
Discriminant Analysis (LSDA) \cite{cai2007locality} and Local Fisher
Discriminant Analysis (LFDA) \cite{sugiyama2007dimensionality},
which are the most related counterparts of MPDA; 3) Tangent space
based methods Linear Local Tangent Space Alignment (LLTSA)
\cite{zhang2007linear} and linearized PFE (we call it LPFE) which
are the linear variations of LTSA and PFE, respectively; 4) Other
types of supervised dimensionality reduction methods Least-Squares
Dimension Reduction (LSDR) \cite{suzuki2013sufficient}. Seven
real-word data sets are used including COIL20, COIL100
\cite{coilDateset}, Face Detection \cite{Alvira01MITface}, a subset
of MNIST \cite{lecun1998mnist} containing the first 2k training and
test images, and three UCI data sets including OptDigits, Semeion
Handwritten and Vehicle \cite{Bache+Lichman:2013}. The configuration
of each data set is shown in Table \ref{table:config}.

The parameters $k$, $\alpha$ and $\gamma$ for both PMPDA and MPDA
are determined by 4-fold cross validation, and the parameters $k'$
and $M$ for the partition algorithm in MPDA are set to be $k'=6$ and
$M=10$ empirically. Furthermore, all the parameters for MFA, LSDA,
LFDA, LLTSA, and LPFE are selected by 4-fold cross validation. The
measure for each round of cross validation is the classification
accuracy on the validation set. Specifically, after training
different dimensionality reduction algorithms on the training set,
we first perform dimensionality reduction on both the training and
validation sets, and then train a classifier using the training set
in the discovered subspace. Finally, by classifying data in the
validation set, we can determine the values of parameters according
to the classification results.  Originally, LPFE is an unsupervised
method. For a fair comparison, LPFE is performed based on a
supervised graph which is identical to the within-class graph $G$
used in MPDA. For each data set, we randomly split certain rates of
data as the training set to compute the subspace, and then classify
the rest of data by the nearest neighbor classifier (1-NN) in the
discovered subspace. Every experimental result is obtained from the
average over 20 splits. For computational efficiency, we use PCA to
preserve 95\% energy for the data sets whose dimensionality are
larger than 100. In addition, we also compare the baseline method
that just employs the 1-NN classifier in the original space without
performing dimensionality reduction.

\begin{figure*}
    \centering
    \subfigure[COIL20\label{figPvD.COIL20}] {
        \includegraphics[width=110pt]{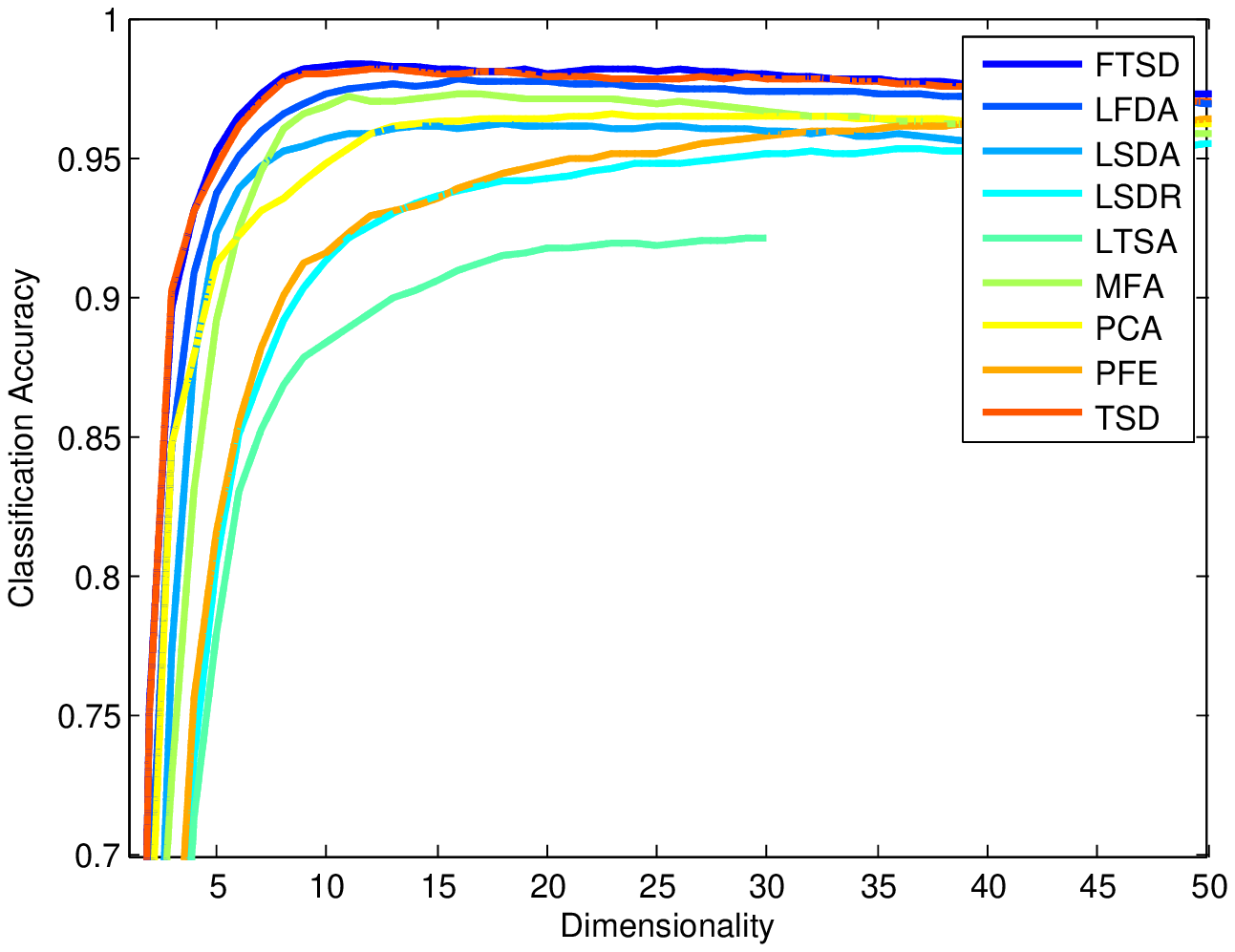}
    }
    \subfigure[COIL100\label{figPvD.COIL100}] {
        \includegraphics[width=110pt]{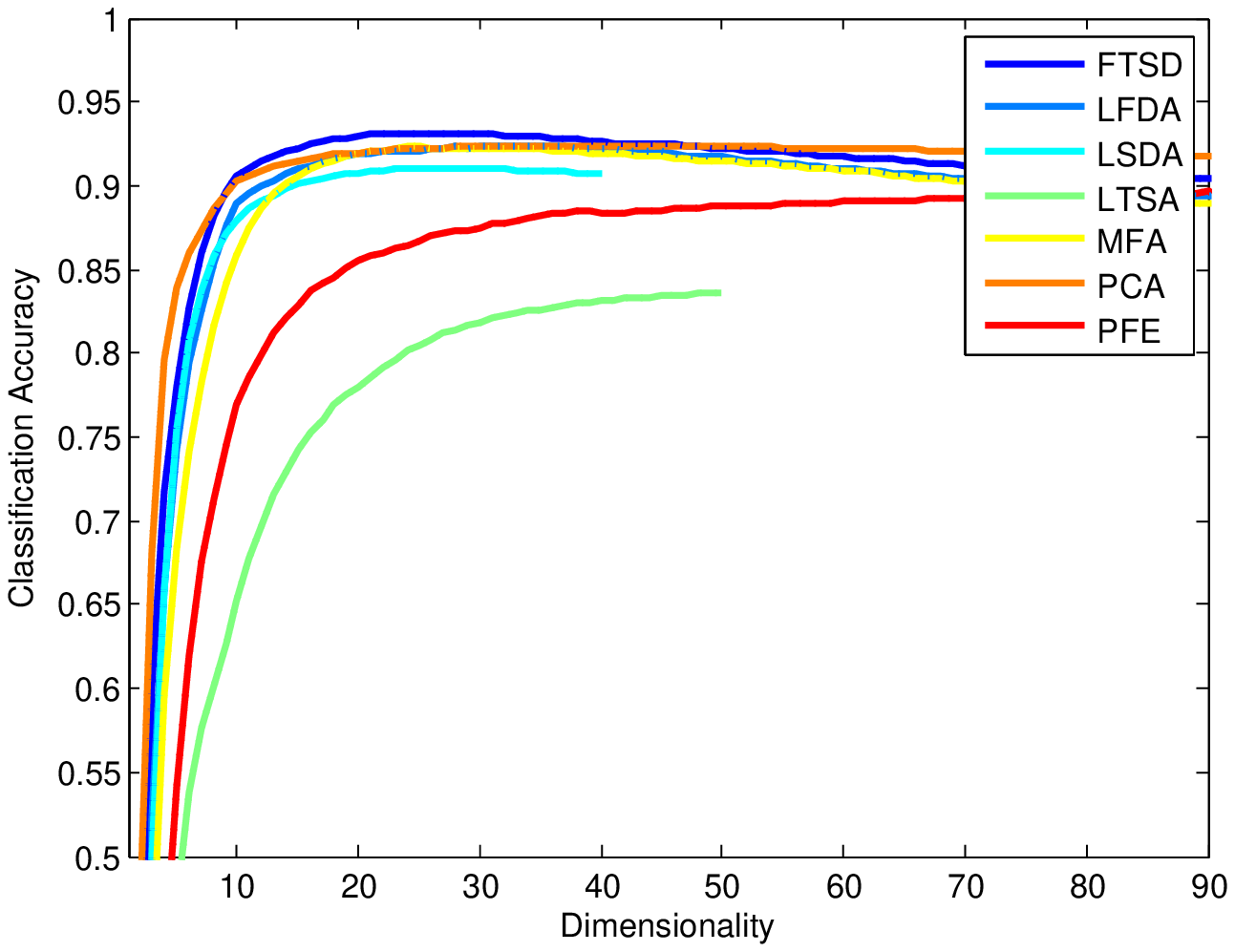}
    }
    \subfigure[Face Detection\label{figPvD.FaceDetection}] {
        \includegraphics[width=110pt]{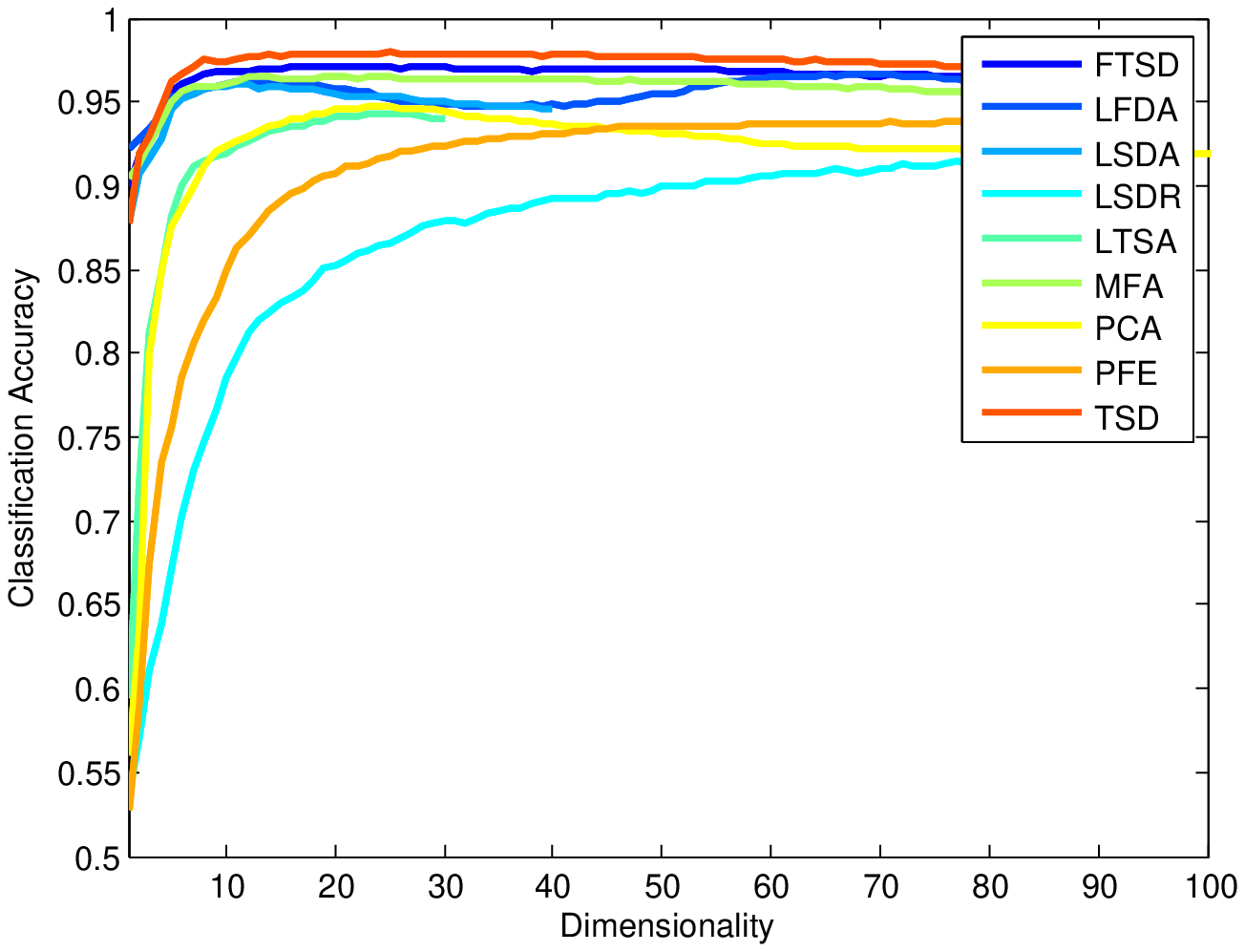}
    }
    \subfigure[MNIST\label{figPvD.MNIST}] {
        \includegraphics[width=110pt]{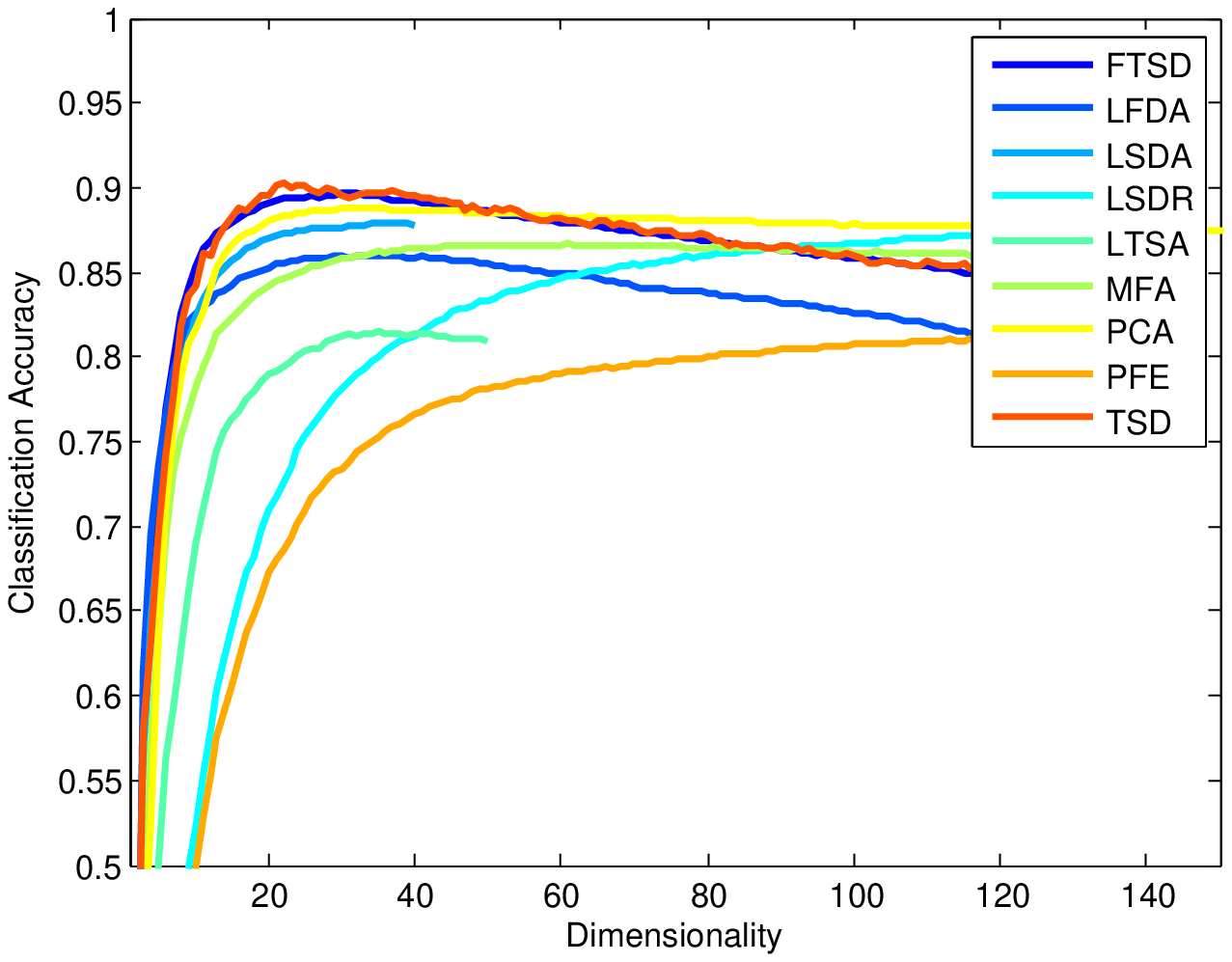}
    }
    \subfigure[OptDigits\label{figPvD.OptDigits}] {
        \includegraphics[width=110pt]{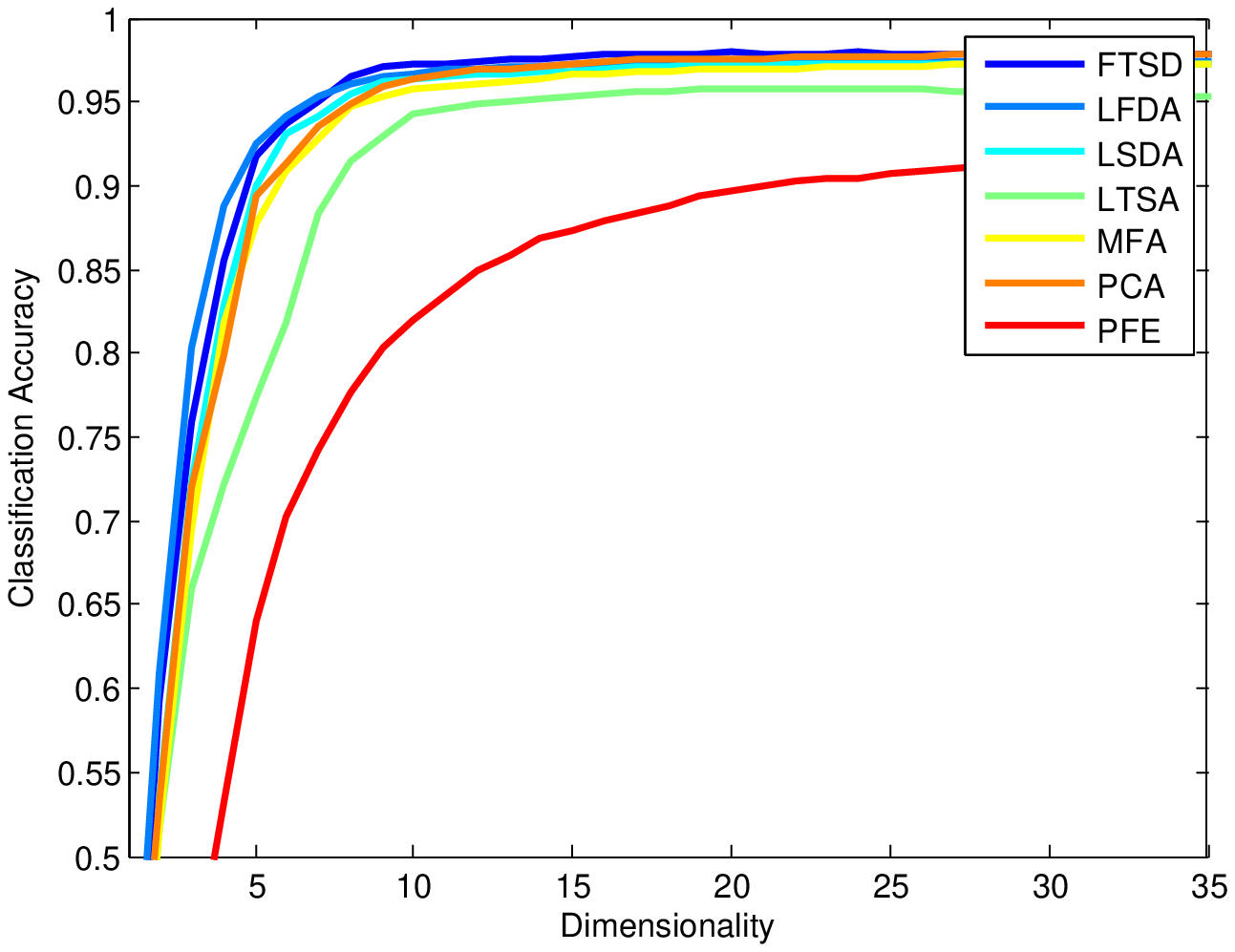}
    }
    \subfigure[Semeion\label{figPvD.Semeion}] {
        \includegraphics[width=110pt]{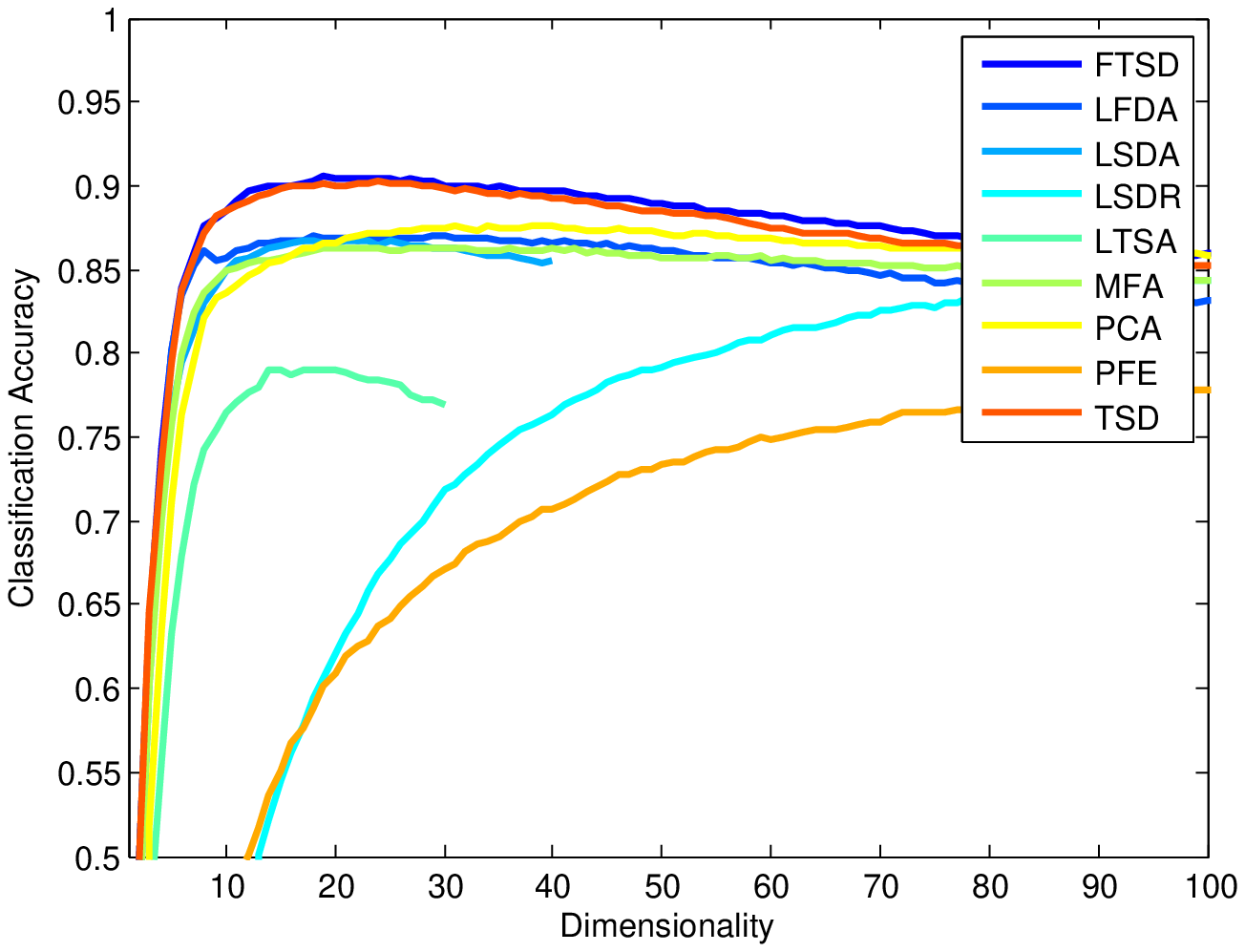}
    }
    \subfigure[Vehicle\label{figPvD.Veicle}] {
        \includegraphics[width=110pt]{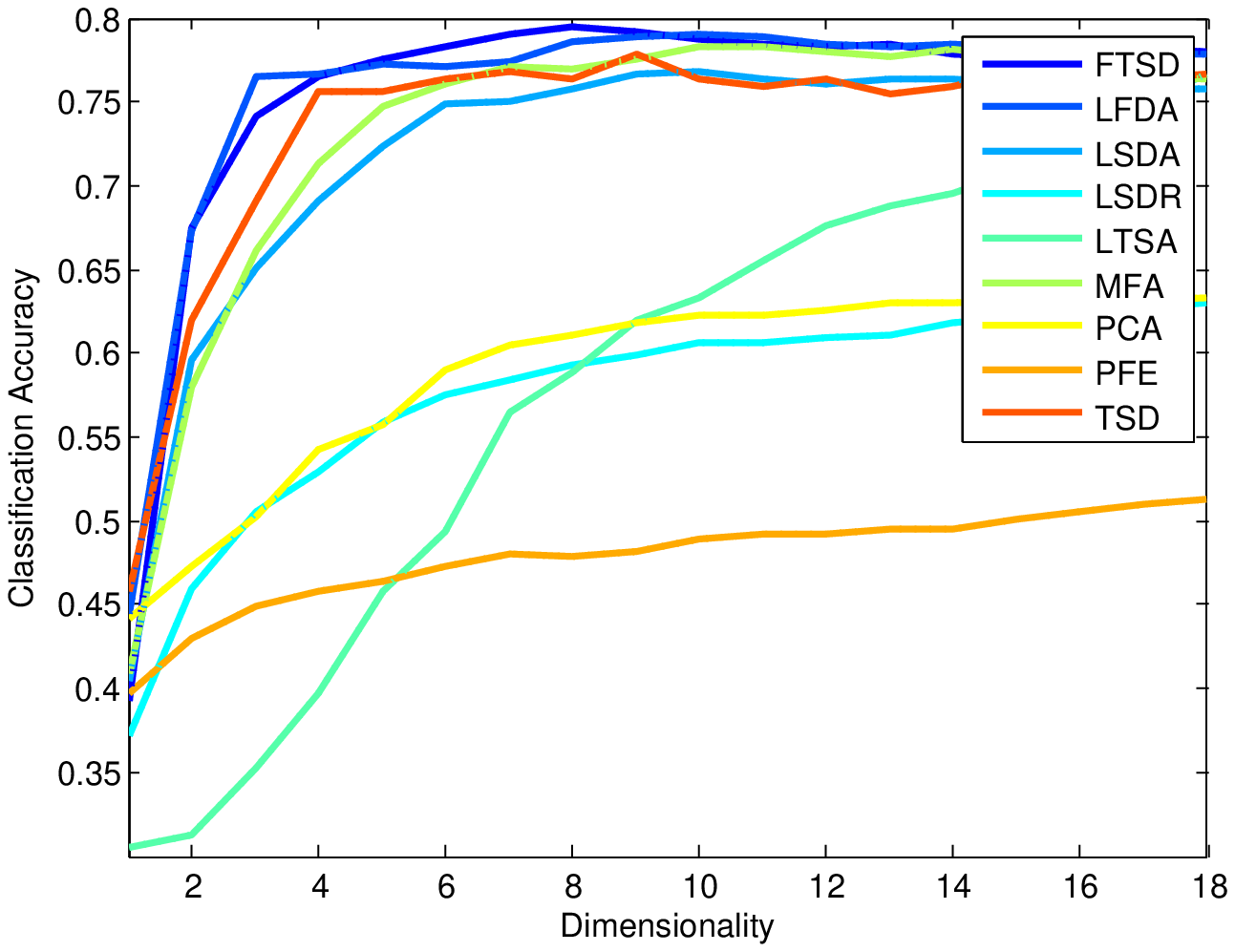}
    }
    \caption{Classification accuracy versus embedding dimensionality
        on different data sets (better viewed in color).\label{figPvD}}
\end{figure*}

Generally, the classification performance varies with the
dimensionality of the subspace. For each method, the best
performance as well as the corresponding dimensionality of the
subspace are reported. PMPDA is not tested on the COIL100 and
OptDigits data sets because of out of memory. LSDR is not tested on
the COIL100 and OptDigits data sets since the execution time is too
long.
Table \ref{table:performance} shows the average error rates of
each method with corresponding dimensionality on different data sets, where
the best method and the comparable one based on
Student's t-test with a p-value of 0.05
are highlighted in bold font. We see that PMPDA or MPDA
outperforms other methods in a statistical significant manner
for all the data sets except the Vehicle data set. This
means that compared with the graph Laplacian based
methods, our methods improve the performance of supervised
dimensionality reduction by taking advantage of the
regional consistency from tangent spaces and keeping
in mind that the within-class similarity shall be achieved
along the varying direction of the data manifold.
LPFE fails to get reasonable results,
which is probably because it has no ability to
separate data from different classes, and it
may lose too much (non-linear) information due to
the linearization.
It is worth noting that although MPDA can be viewed as the
approximation of PMPDA, it still obtains comparable or better
results than PMPDA. This suggests that the manifold partition
strategy itself is able to improve the performance of dimensionality
reduction, because it provides MPDA with more flexibility to
estimate tangent spaces. In addition, although not shown in Table
\ref{table:performance}, if we remove PMPDA out of the comparison,
MPDA becomes the best method based on t-test with p=0.05. This
demonstrates that MPDA is consistently better than its counterparts.
Figure \ref{figPvD} depicts how the mean classification accuracy
varies with respect to the dimensionality of embedding spaces on
different data sets. It shows that MPDA and PMPDA work quite well.
Particularly, except for the Vehicle data set, MPDA and PMPDA (if
applicable) consistently obtain the best results with respect to the
dimensionality of embedding spaces. To further evaluate the
effectiveness of MPDA, we also conduct experiments on the data sets
that have been tested by the authors of its counterparts. In this
case, results from the existing algorithms can be cited from the
corresponding original paper for fairer comparisons. According to
Table \ref{table:performance}, LFDA seems to be the best algorithm
except for MPDR and PMPDR. Because of this, we focus on comparing
MPDA with LFDA on the USPS handwritten digit data set according to
the configuration of LFDA's original paper. Again, MPDR and PMPDR
outperform their counterparts with statistical significance. Please
refer to the supplementary material for details.

\begin{figure*}
    \centering
    \subfigure[$k \in \{1,2,\ldots,15\}$\label{figPara.k}] {
        \includegraphics[width=110pt]{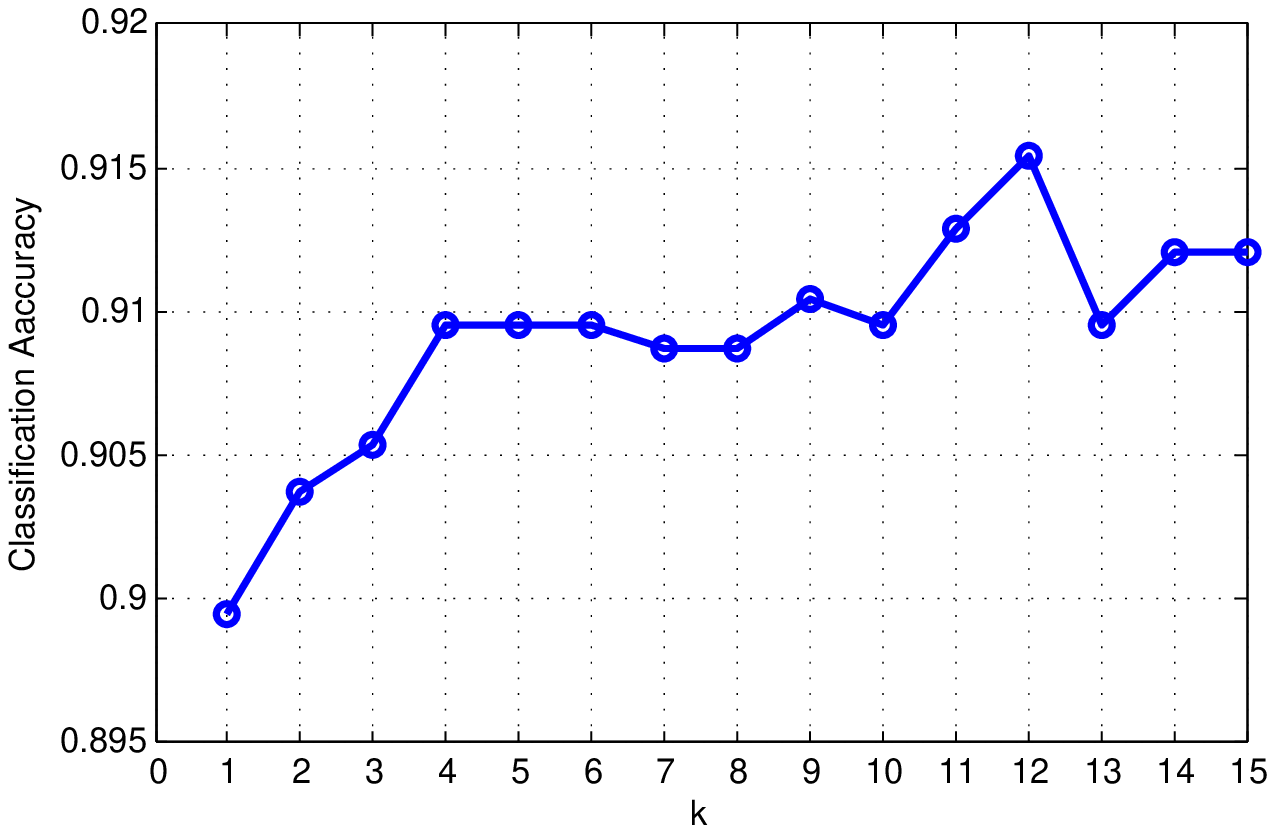}
    }
    \subfigure[{\footnotesize $\gamma \in \{10^{-5},10^{-4},\ldots,10^{5}\}$}\label{figPara.gamma}] {
        \includegraphics[width=110pt]{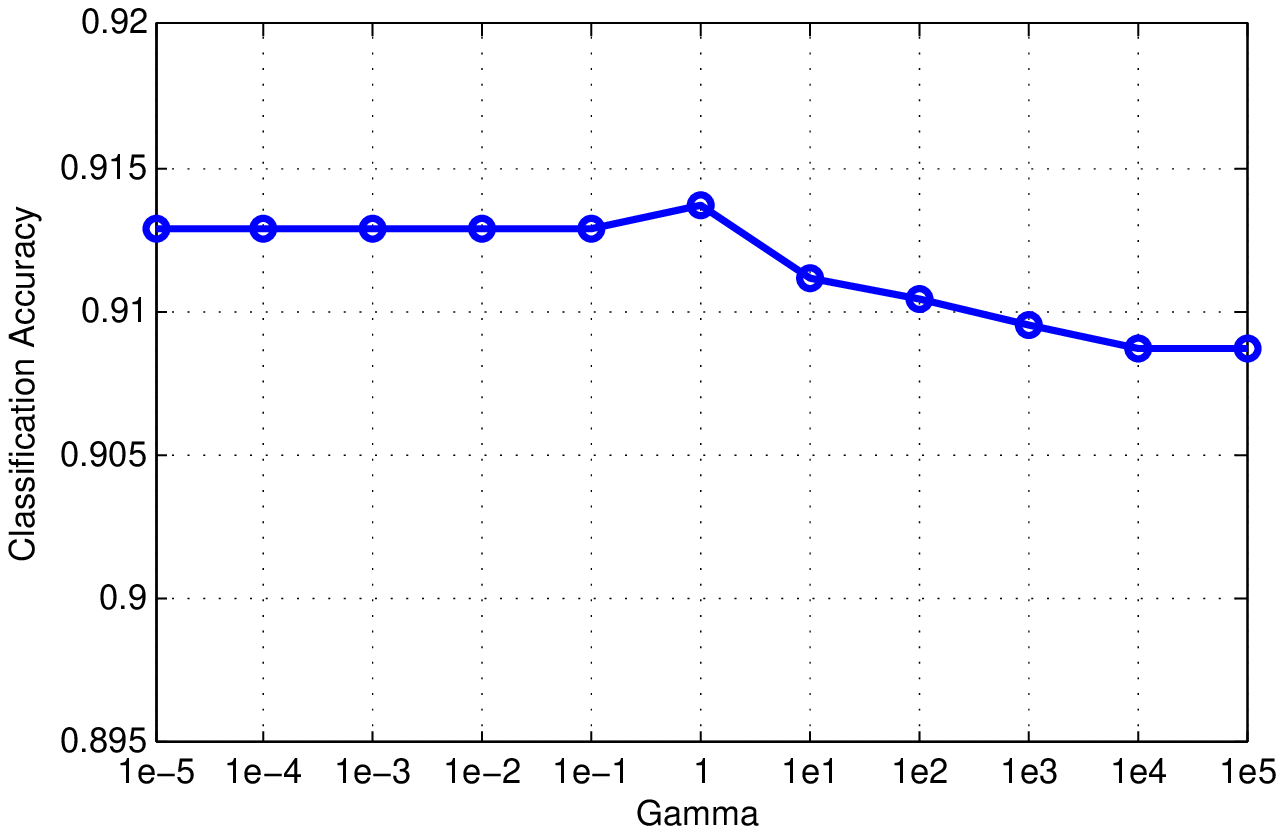}
    }
    \subfigure[$M \in \{5,10,\ldots,50\}$\label{figPara.M}] {
        \includegraphics[width=110pt]{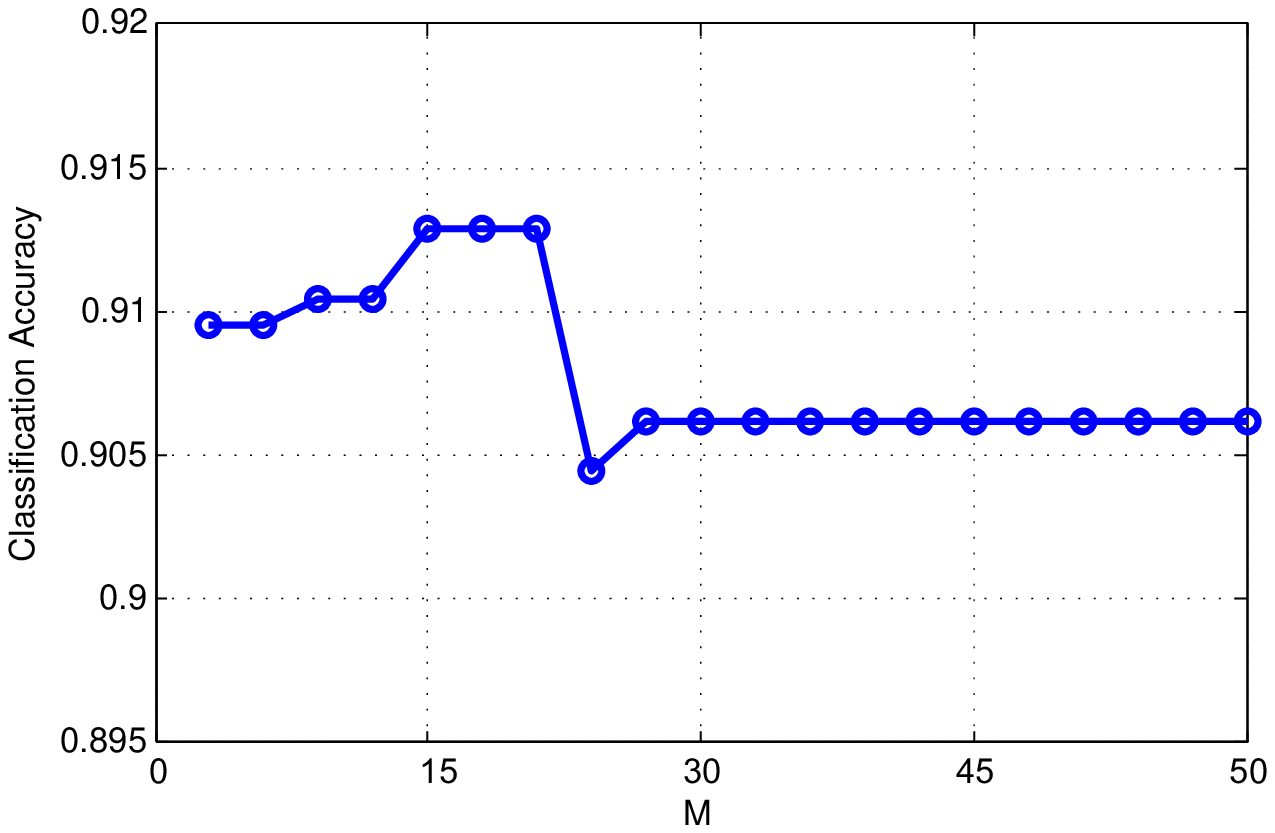}
    }
    \subfigure[$k' \in \{1,2,\ldots,15\}$\label{figPara.k'}] {
        \includegraphics[width=110pt]{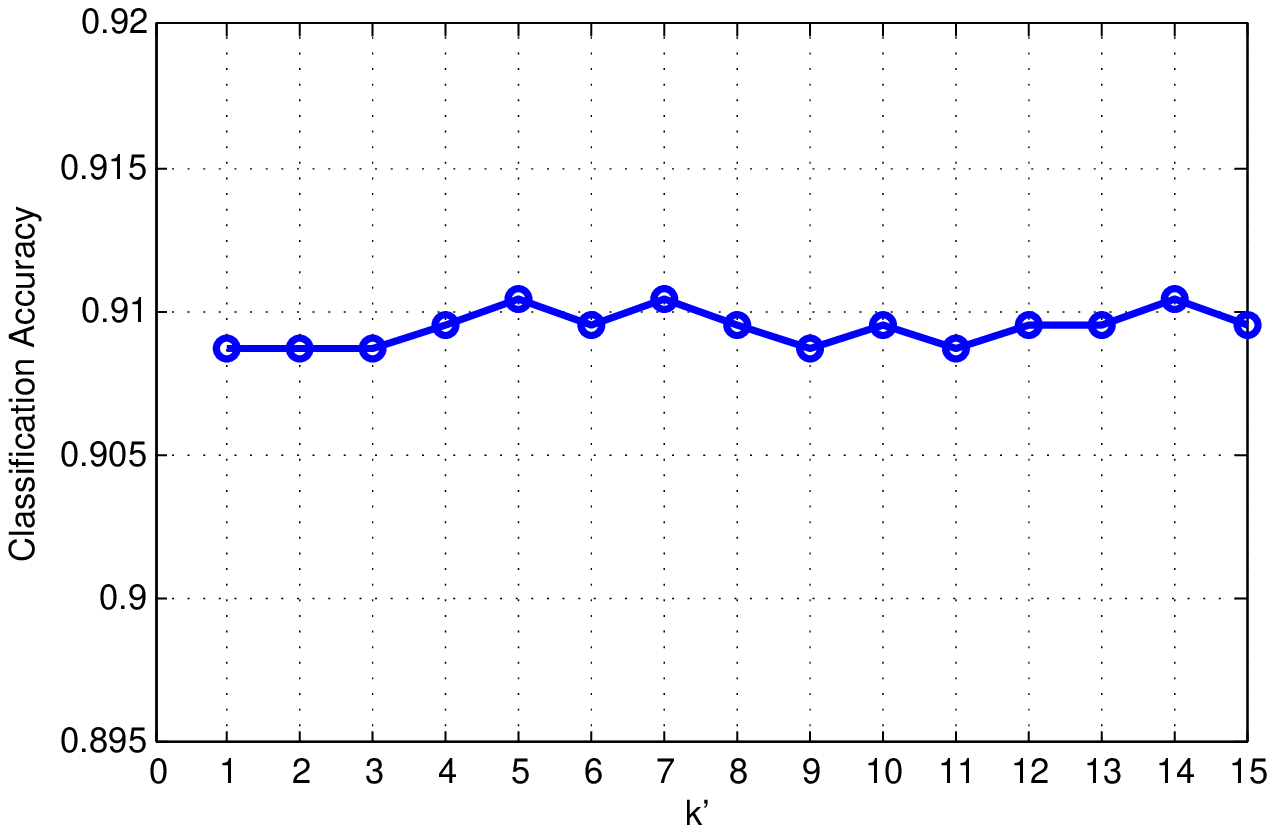}
    }
    \caption{Average classification accuracy of MPDA with respect to the values of
        different parameters on the Semeion data set.\label{figPara}}
\end{figure*}

\subsection{Parameter Sensitivity}
In this section, we evaluate the parameter sensitivity of MPDA on
the Semeion Handwritten data set. Specifically, we aim to test how
the performance of MPDA varies with its parameters $k$, $\gamma$,
$M$, and $k'$, respectively. To this end, the default values of $k$,
$\gamma$, $M$ and $k'$ are set to be $5$, $1$, $10$ and $6$,
respectively. And we alternately change one of these parameters to
evaluate the performance of MPDA when the other parameters are
fixed. Figure \ref{figPara} implies that $k$ and $M$ are more
important than $\gamma$ and $k'$. Their values should be determined
properly, while those of $\gamma$ and $k'$ seem to have no
significant influence on the performance of MPDA. Overall, MPDA get
stable results as its parameters change, where the classification
accuracy ranges from $90\%$ to $92\%$. Therefore, MPDA is relatively
insensitive to the changes of parameters.

\section{Conclusion} \label{sec:con}
In this paper we have proposed a tangent space based linear
dimensionality reduction method named Manifold Partition
Discriminant Analysis (MPDA). By considering both pairwise
differences and piecewise regional consistency, MPDA can find a
linear embedding space where the within-class similarity is achieved
along the direction that is consistent with the local variation of
the data manifold, while nearby data belonging to different classes
are well separated. Different to graph Laplacian methods that
capture only the pairwise interaction between data points, our
method capture both pairwise as well as higher order interactions
(using regional consistency) between data points.

As a crucial part of MPDA, the manifold partition
strategy plays a key role in
preserving the manifold structure
to improve the measure of the within-class similarity.
It not only enables MPDA to adaptively determine the number of
dimensions of each linear subspace, but also can be adopted by
other tangent space base methods to make them more efficient.
The experiments on multiple real-world data sets
have shown that compared with existing works MPDA
can obtain better classification results.

\ifCLASSOPTIONcompsoc
  \section*{Acknowledgments}
\else
  \section*{Acknowledgment}
\fi

This work is supported by the National Natural Science Foundation of China under Projects 61370175, and Shanghai Knowledge Service Platform Project (No. ZF1213).

\ifCLASSOPTIONcaptionsoff
  \newpage
\fi



\bibliographystyle{IEEEtran}
\bibliography{MPDA}
\begin{IEEEbiography}[{\includegraphics[width=1in,height=1.7in,clip,keepaspectratio]{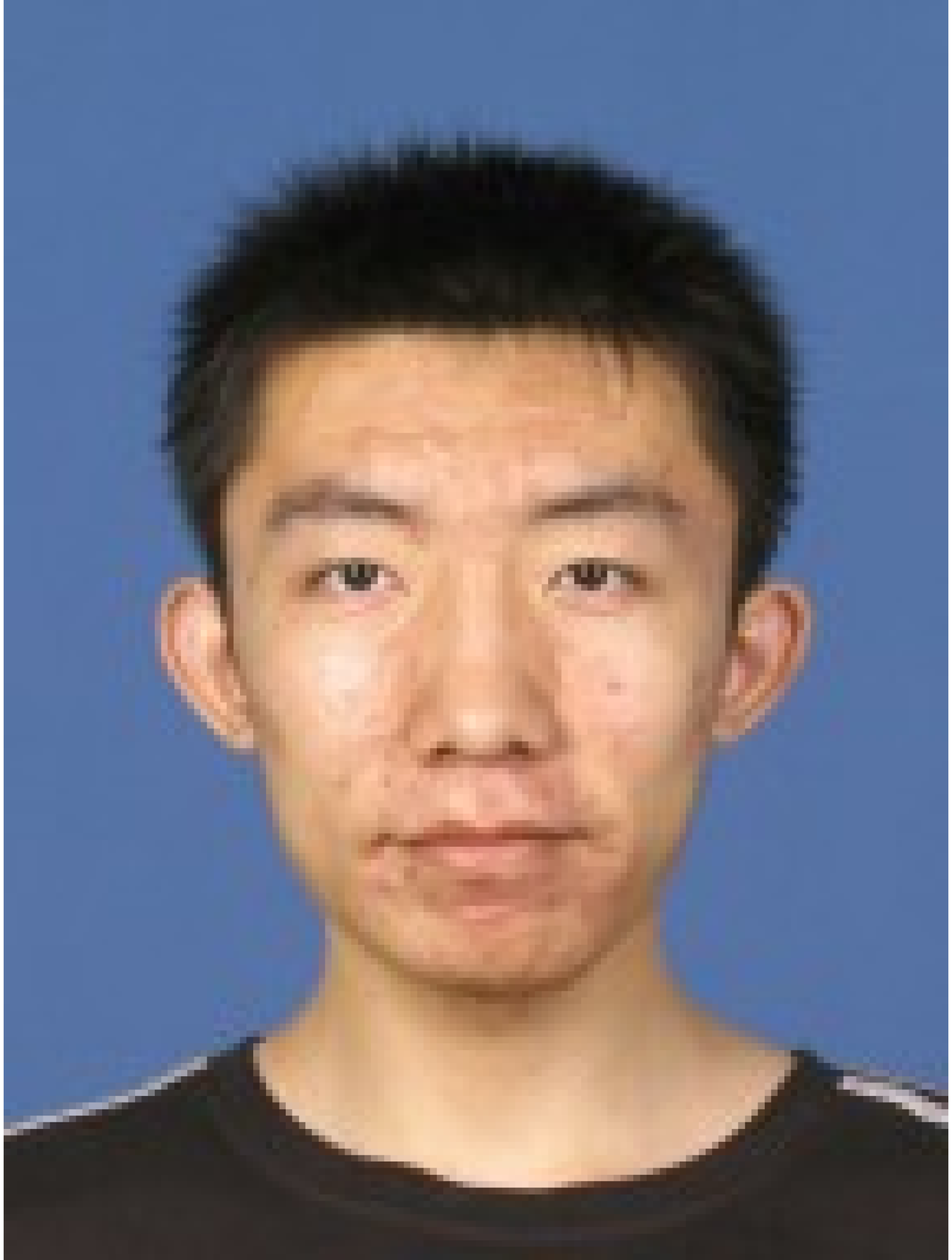}}]{Yang Zhou}
is a master student in the Pattern Recognition and Machine Learning Research Group, Department of Computer Science and Technology, East China
Normal University. His research interests include
pattern recognition, manifold learning, dimensionality reduction, etc.
\end{IEEEbiography}

\begin{IEEEbiography}[{\includegraphics[width=2in,height=2.7in,clip,keepaspectratio]{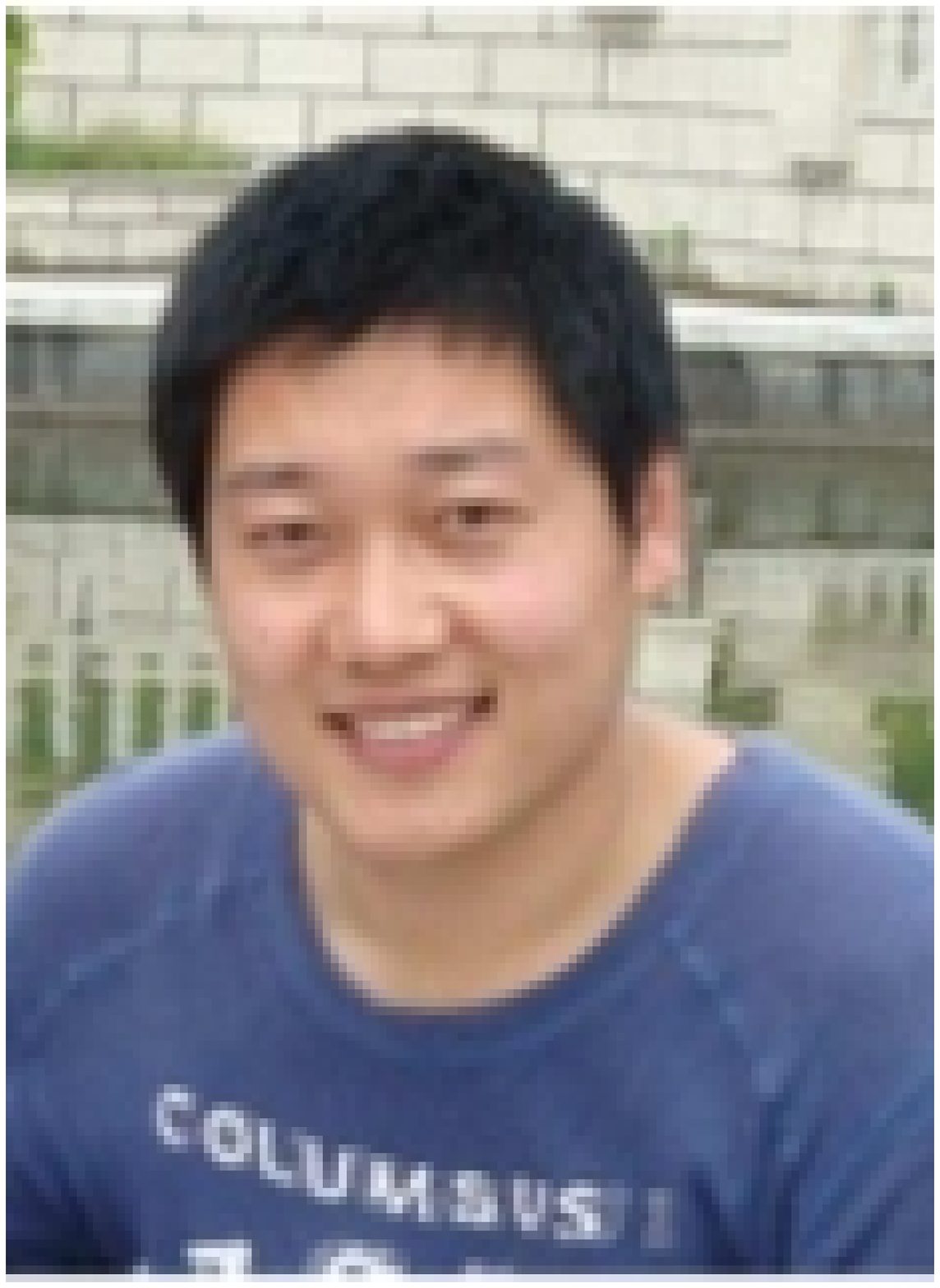}}]{Shiliang Sun}
is a professor at the Department of Computer Science and Technology
and the head of the Pattern Recognition and Machine Learning
Research Group, East China Normal University. He received the Ph.D.
degree in pattern recognition and intelligent systems from the
Department of Automation and the State Key Laboratory of Intelligent
Technology and Systems, Tsinghua University, Beijing, China, in
2007. From 2009 to 2010, he was a visiting researcher at the
Department of Computer Science, University College London, working
within the Centre for Computational Statistics and Machine Learning.
In July 2014, he was a visiting researcher at the Department of
Electrical Engineering, Columbia University, New York. He is on the
editorial boards of multiple international journals including
Neurocomputing and IEEE Transactions on Intelligent Transportation
Systems. His research interests include kernel methods, learning
theory, multi-view learning, approximate inference, sequential
modeling and their applications, etc.

\end{IEEEbiography}

\end{document}